\documentclass[journal]{IEEEtran}

\usepackage{epsfig}
\usepackage{graphicx}
\usepackage{cite}
\usepackage{amsmath}
\usepackage{amssymb}
\usepackage{bm}
\usepackage{algorithm}
\usepackage{algorithmicx}
\usepackage{algpseudocode}
\usepackage{booktabs}
\usepackage{multirow}
\usepackage{graphics}
\usepackage{ulem}
\usepackage{color,soul}
\usepackage{enumerate}
\usepackage[table]{xcolor}
\usepackage{url}
\usepackage{cite}
\usepackage{setspace}

\renewcommand{\algorithmicensure}{\textbf{Output:}}

\begin{document}

\title{Denoising Hyperspectral Image with Non-i.i.d. Noise Structure}


\author{Yang Chen, Xiangyong Cao, Qian Zhao, Deyu Meng,~\IEEEmembership{Member,~IEEE,} Zongben Xu
\thanks{Yang Chen, Xiangyong Cao, Qian Zhao, Deyu Meng (corresponding author), Zongben Xu are with the School of Mathematics and Statistics and Ministry
	of Education Key Lab of Intelligent Networks and Network
	Security, Xi'an Jiaotong University, Xi'an 710049, China (e-mail: dymeng@mail.xjtu.edu.cn).}}

%
%

\markboth{Journal of \LaTeX\ Class Files,~Vol.~14, No.~8, August~2015}%
{Shell \MakeLowercase{\textit{et al.}}: Bare Demo of IEEEtran.cls for IEEE Journals}
%


\maketitle

\begin{abstract}
Hyperspectral image (HSI) denoising has been attracting much research attention in remote sensing area due to its importance in improving the HSI qualities. The existing HSI denoising methods mainly focus on specific spectral and spatial prior knowledge in HSIs, and share a common underlying assumption that the embedded noise in HSI is independent and identically distributed (i.i.d.). In real scenarios, however, the noise existed in a natural HSI is always with much more complicated non-i.i.d. statistical structures and the under-estimation to this noise complexity often tends to evidently degenerate the robustness of current methods. To alleviate this issue, this paper attempts the first effort to model the HSI noise using a non-i.i.d. mixture of Gaussians (NMoG) noise assumption, which is finely in accordance with the noise characteristics possessed by a natural HSI and thus is  capable of adapting various noise shapes encountered in real applications. Then we integrate such noise modeling strategy into the low-rank matrix factorization (LRMF) model and propose a NMoG-LRMF model in the Bayesian framework. A variational Bayes algorithm is designed to infer the posterior of the proposed model. All involved	parameters can be recursively updated in closed-form. Compared with the current techniques, the proposed method performs more robust beyond the state-of-the-arts, as substantiated by our experiments implemented on synthetic and real noisy HSIs.
\end{abstract}

\begin{IEEEkeywords}
Hyperspectral image denoising, Non-i.i.d. Noise modeling, Low-rank matrix factorization.
\end{IEEEkeywords}

%
\IEEEpeerreviewmaketitle

\section{Introduction}

\IEEEPARstart{H}{yperspectral} image (HSI) is captured from sensors over various bands and contains abundant spatial and spectral knowledge across all these bands beyond the traditional gray-scale or RGB images. Due to its preservation of full-bands information under a real scene, it has been widely used in military and civilian aspects such as terrain detection, mineral exploration, pharmaceutical counterfeiting, vegetation and environmental monitoring~\cite{plaza2009recent,goetz2009three,zhao2008object,chen2011hyperspectral,yuen2010introduction,qian2013hyperspectral}.   In real cases, however, a HSI is always corrupted by noises due to equipment limitations, such as sensor sensitivity, photon effects and calibration error~\cite{skauli2011sensor}. Besides, due to the limited radiance energy and generally narrow band width, the energy captured by each sensor might be low and thus the shot noise and thermal noise then always happen inevitably. These noise severely degrades the quality of the imagery and limits the performance of the subsequent processing, i.e., classification~\cite{sathya2015hyperspectral}, unmixing~\cite{qian2011hyperspectral, bioucas2012hyperspectral} and target detection~\cite{stein2002anomaly}, on data. Therefore, it is a critical preprocessing step to reduce the HSI noise ~\cite{peng2014decomposable, he2016total} to a general HSI processing task.

The simplest denoising way is to utilize the traditional 2-D or 1-D denoising methods to reduce noise in the HSI  pixel by pixel \cite{green1988transformation} or band by band  \cite{elad2006image,dabov2007image}, \cite{liu2016weighted,shao2014heuristic,yan2013nonlocal}.  However, this kind of processing method ignores the correlations between different spectral bands or adjacent pixels, and often results in a relatively low-quality result. Recently, more efficient methods have been proposed to handle this issue.  the main idea is to elaborately encode the prior knowledge on the structure underlying a natural HSI, especially the characteristic across the spatial and spectral dimensionality. E.g., Othman and Qian ~\cite{othman2006noise} made an initial attempt to this issue by designing a hybrid spatial-spectral derivative-domain wavelet shrinkage model based on the dissimilarity of the signal regularity existing along the space and spectrum of a natural HSI. And then Chen et al. ~\cite{chen2008simultaneous} proposed another approach to encoding both spatial and spectral knowledge  by combining the  bivariate wavelet thresholding with principal component analysis. To further enhance the denoising capability, Yuan et al.~\cite{yuan2012hyperspectral}  employed a spectral-spatial adaptive total variation. Later, Chen et al.~\cite{chen2012hyperspectral} proposed a spatially adaptive weighted prior by combining the smoothness and discontinuity preserving properties along the spectral domain. By further considering the spatial and spectral dependencies, Zhong and Wang~\cite{zhong2013multiple} proposed a multiple-spectral-band CRF (MSB-CRF) model in a unified probabilistic framework.

Besides, by explicitly treating HSI data as a tensor, a series of methods that expanding wavelet-based method from 2-D to 3-D has been proposed. E.g., Dabov \cite{dabov2007video} designed VBM3D method by applying the concepts of grouping and collaborative filtering to video denoising. Then, Letexier et al.  \cite{letexier2008noise} proposed a generalized multi-dimensional Wiener filter for denoising hyperspectral image. Similarly, Chen et al.\cite{chen2011denoising} extended Sendur and Selesnick's bivariate wavelet thresholding from 2-D image denoising to 3-D data cube denoising.  For better denoising results, Chen et al.~\cite{chen2012signal} proposed a new signal denoising algorithm  by using neighbouring wavelet coefficients, which considered both translation-invariant (TI) and non-TI versions.  Later, as an extension of BM3D method, Maggioni et al.~\cite{BM4D2013TIP} presented BM4D model.
Meanwhile, another type of method that based on tensor decomposition  has appeared.  Karami et al.~\cite{karami2011noise}  developed a Genetic Kernel Tucker Decomposition (GKTD) algorithm to exploit both spectral and the spatial information in HSIs. To address the uniqueness of multiple ranks of Tucker Decomposition, Liu et al. \cite{liu2012denoising} proposed PARAFAC method that make  the number of estimated rank reduced to one. Later, Peng et al.~\cite{peng2014decomposable} proposed a decomposable nonlocal tensor dictionary learning (TDL) model, which fully considered the non-local similarity over space and the global correlation across spectrum. Among these methods, BM4D and TDL achieved the state-of-the-art performance in more general noisy MSI cases.

Most of current HSI denoising methods have mainly considered the HSI prior spectral and spatial knowledge into their methods, while only use $L_2$ loss term to rectify the deviation between the reconstruction and the original HSI. In the viewpoint of statistical theory, the utilization of such loss term implies that the HSI data noise follows an i.i.d. Gaussian distribution. However, in real scenarios, HSI noises generally have more complicated statistical structures. This means that such easy loss term is too subjective to reflect the real cases and inclines to degenerate the performance of the methods in more complex while more realistic non-i.i.d. noise case .
%

The idea of considering more complex HSI noise beyond only Gaussian in HSI denoising has been attracting attention very recently in the framework of low-rank matrix analysis. Since adjacent HSI bands usually exhibit strong correlations, by reshaping each band as a vector and stacking all the vectors into a matrix, the reshaped HSI matrix can be assumed to be with low rank. Various low-rank matrix models have been presented in different noise scenes in recent decades. Along this line, the classical Low-rank Matrix Factorization (LRMF) model is presented by K.Mitra et al.~\cite{okatani2011efficient} and T.Okatani et al.~\cite{mitra2010large} for Gaussian noise, and its global optimal solution can be directly obtained by using Singular Value Decomposition (SVD)~\cite{haardt1996method}. To add more robustness, the $L_ 1$-norm  LRMF~\cite{ding2006r,eriksson2010efficient,ji2010robust,ke2005robust} is generally used and many algorithms have been designed to solve this model, such as $L_1$ Wiberg~\cite{eriksson2010efficient}, CWM~\cite{meng2013cyclic} and RegL1ALM~\cite{zheng2012practical}. This $L_1$-norm LRMF model actually assumes an i.i.d. Laplacian noises embedded in data. To handle more complex noise cases, Meng and De la Torre~\cite{meng2013robust} modeled the noise as more adaptable i.i.d. mixture of Gaussians (MoG) distributions to represent noise. Such noise modeling idea was futher extended to the Bayesian framework by Chen et al.~\cite{chen2015bayesian}, to RPCA by Zhao et al.~\cite{zhao2014robust} and  to the tensor factorization by Chen et al.~\cite{han2016CVPR}  . Very recently, Cao et al. ~\cite{cao2015PMoEP,cao2016PMoEP} modeled the noise as a more general i.i.d. mixture of Exponential Power (MoEP) distribution, which achieves competitive performance in HSI denoising under real noise scenarios. Besides, some other attempts have also been proposed to model the noise as a combination of sparse and Gaussian noise~\cite{babacan2012sparse}, and Zhang et al.~\cite{zhang2014hyperspectral} also utilized this idea for HSI denoising task. Later, He et al.~\cite{hetotal} further enhance the capability of the method by adding a TV regularization to low-rank reconstruction. Besides, based on the mixture noise assumption, He et.al ~\cite{he2015hyperspectral} proposed a noise-adjusted low-rank methods and Wang et al.~\cite{wangdenoising} proposed a GLRR denoising method for the reconstruction of the corrupted HSIs. All these approaches also achieve good performance on HSI denoising in mixed noise cases.

From Gaussian noise assumption to MoEP noise assumption, such advancements make the model more adaptive to various HSI noises encountered in practice.  However, all of the aforementioned LRMF models just simply assume that the HSI noise is i.i.d., which is more or less deviated from the practical scenarios, where the noises in a HSI is generally with non-i.i.d. configurations. In this sense, there is still a large room to further improve the capability of current HSI denoising techniques especially under real complicated scenarios.

\begin{figure}[t]
	\centering
	\includegraphics[width=\linewidth]{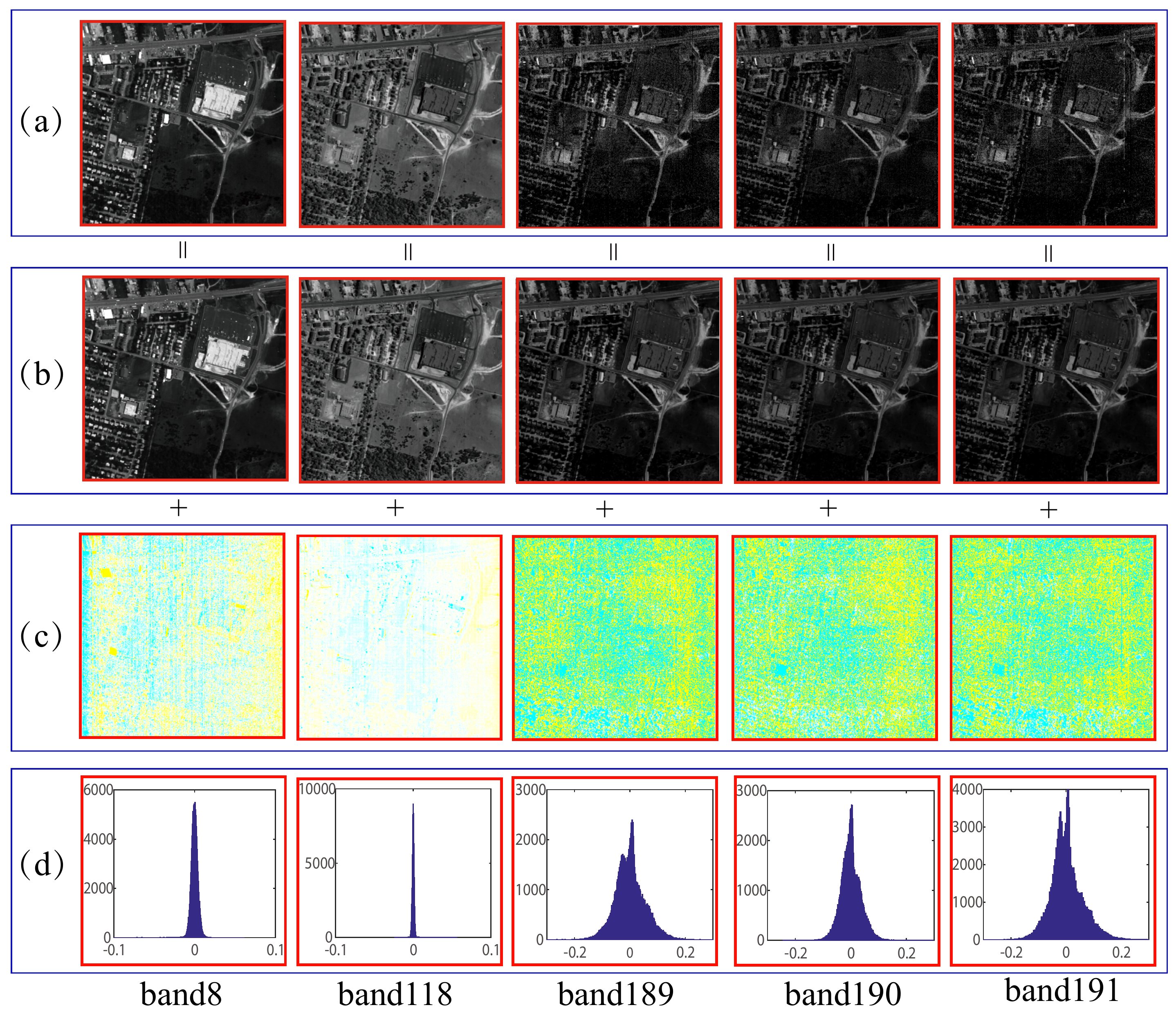}
\vspace{-6mm}
	\caption{(a) Different bands of images in the original HSI. (b) Restored HSIs obtained by our proposed method. (c) Extracted noise by the proposed method. (d) Histograms of the noise in all bands. \label{fig1} }
\end{figure}

To make this point clearer, let's try to analyze some evident noise characteristics possessed by a HSI collected from real cases. Fig. \ref{fig1} presents a real HSI case for auxiliary understanding. Firstly, an image in a band is generally collected under the similar sensor setting, and thus the i.i.d. distribution is a rational assumption for the noise over the elements in the band. This can be evidently observed from the band-noise of the HSI image shown in Fig. \ref{fig1}. Secondly, due to the difference of the sensor sensitivity for collecting images located in various HSI bands, the noise of different-band-images always depicts evident distinctions~\cite{he2015hyperspectral}. From Fig. \ref{fig1}, it is easy to see that images located in some bands are very clean, while are extremely noisy in some others~\cite{zelinski2006denoising,goetz2009three}. Thirdly, albeit different, the noise distributions along different bands have certain correlation. E.g., along adjacent bands, the noise tends to be more or less similar since images on neighboring spectrums are generally collected under small deviation of sensor settings and wavelength ~\cite{zhong2013multiple}. From Fig. \ref{fig1}, it is easy to see the noise similarity for images located in 189-191 bands. Accordingly, the real HSI noise distribution is generally non-i.i.d. and has a more complicated configurations than the i.i.d. noise assumption of the current HSI denoising techniques. Such deviation inclines to make their performance degenerate under more practical cases, which will be clearly observed in our experiments.

To address this issue, this paper proposes a new noise modeling framework, by carefully designing noise distribution structure to make it possibly faithfully deliver the real HSI noise configurations. Specifically, we model the noise of each HSI band with different Mixture of Gaussian (MoG) distributions (i.e., parameters of MoG are different). Besides, MoG parameters across different bands are encoded under a similar top-level prior distribution, representing the correlation between noise distributions across different bands. In this way, the non-i.i.d. noise structure under a practical HSI image can be more properly encoded and an appropriate HSI denoising effect is thus to be expected.

In this study, we embed such noise modeling framework into the low-rank matrix factorization (LRMF) realization, which easily assumes a low-rank structure (both spatial  and spectral) of the to-be-recovered HSI. Our experimental results show that, even under this simple setting of HSI structure, such noise modeling idea can greatly help enhance the HSI denoising ability beyond previous techniques and  obtain the state-of-the-art performance in real HSI denoising tasks.

The rest of the paper is organized as follows: The proposed model and the corresponding variational inference algorithm are presented in Section 2. Experimental results are shown in Section 3. Finally, conclusions are drawn in Section 4. Throughout the paper, we denote scalars, vectors, matrices, tensors as non-bold letters, bold lower case letters,  bold upper case letters and decorated letters, respectively.

\section{Non-i.i.d. MoG method for HSI denoising}

In this section, we first introduce our non-i.i.d. noise encoding strategy and then propose a non-i.i.d. MoG LRMF (NMoG-LRMF) model by using this strategy in the LRMF model for HSI denoising. Finally, the corresponding variational inference algorithm for the proposed model is designed.

\subsection{Non-i.i.d. noise encoding}

Let's represent a given HSI as a matrix $ \bm{Y}\in \mathbb{R} ^{N\times B} $, where $ N $ and $ B $ mean the spatial and spectral dimensions of the HSI, respectively. By assuming that noise is addictive, we have the following expression:
\begin{equation}
\bm{Y = L+E},
\end{equation}
where $ \bm{L} \in \mathbb{R}^{ N\times B } $ denotes the clean HSI  and $ \bm{E} \in \mathbb{R}^{ N\times B } $ denotes the embedded noise.

In the previous section, we have summarized three intrinsic properties possessed by HSI noise: (i) Noise of an image located in the same band are  i.i.d.; (ii) Noise of images located in different bands are non-identical; (iii) Noise distributions in different bands are correlated.

Based on the above properties (i) and (ii) of HSI noise structure, we model noise located in each band as an independent MoG distribution while assume that the parameters for MoG distributions in different bands are different. The MoG is utilized due to its universal approximation capability for any continuous densities~\cite{maz1996approximate}, which has been extensively used and studied in~\cite{meng2013robust,zhao2014robust}.

Denote $ e_{ij}  $ as the element located in $ i^{th} $ row and $ j^{th} $ column of the noise matrix $ \bm{E} $, and as aforementioned, the noise distribution located in the $ j^{th} $ band can then be modeled as:
\begin{equation}\label{np}
p(e_{ij})=\sum_{k=1}^{K}{\pi_{jk} \mathrm{N}(e_{ij}|\mu_{jk},\tau_{jk}^{-1})},
\end{equation}
where $\pi_{jk}$ is the mixing proportion with $\pi_{jk} \geq 0$ and $ \sum_{k=1}^{K}{\pi_{jk}}=1 $, $ K $ is the Gaussian component number, $ \mu_{jk} $ and $ \tau_{jk} $ are mean and precision of the $ k^{th} $ Gaussian $ \mbox{component} $ in the $ j^{th} $ band, respectively. Note that MoG parameters $ \pi_{jk}, \mu_{jk}$
and $ \tau_{jk} $ are different, implying different MoG distributions across different bands.

\begin{figure}[t]
	\centering
	\includegraphics[width=0.8\linewidth]{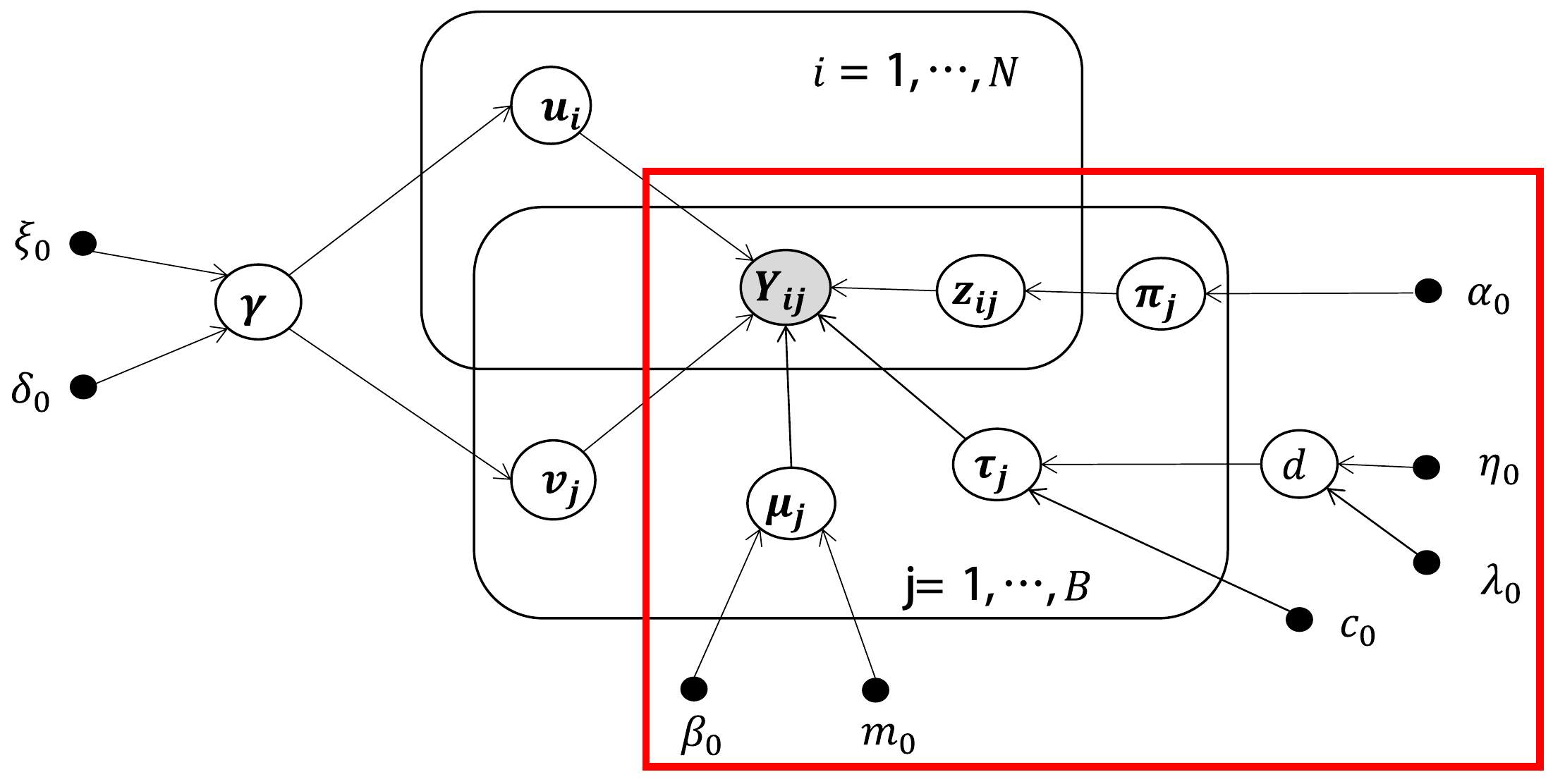}
\vspace{-2mm}
	\caption{Graphical model of NMoG-LRMF. $ Y_{ij} $ denotes the $ i^{th}$ HSI element in its $ j^{th} $ band. $ \bm{u}_{i} $ and $ \bm{v}_{j} $ are columns of low-rank matrix $ U $ and $ V $, respectively, generated from a Gaussian distribution with precise $ \bm{\gamma}$, with Gamma prior distribution with hyper-parameters $\xi_0 $ and $ \delta_0 $. The portion in the red box corresponds to the noise encoding part. $ \bm{\mu}_j $, $ \bm{\tau}_j $ are the $ k $-dimentional vectors representing the mean and variance of all MoG components in the $j^{th}$ band, with hyper-parameters $ \beta_0, m_0 ,d, c_0 $, where $d$ is with hyper-parameters $ \eta_0, \lambda_0 $. $ \bm{z}_{ij} $ is the hidden variable generated from Multinomial distribution with parameter $ \bm{\pi}_{j} $, with hyper-parameters $ \alpha_0 $.} \label{gm}
\end{figure}

Considering the noise property (iii), we further provide the hypothesis that MoG parameters $ \bm{\mu}_js $ and $ \bm{\tau}_js $ of all bands are generated from a two-level prior distribution:
\begin{equation}\label{mutao}
\begin{split}
\mu_{jk} ,\tau_{jk}&\sim \mathrm{N}(\mu_{jk}|m_0,(\beta_0\tau_{jk})^{-1}) \mathrm{Gam}(\tau_{jk}|c_0,d),\\
d & \sim \mathrm{Gam}(d|\eta_0,\lambda_0),
\end{split}
\end{equation}
where $ \mathrm{Gam}(\cdot) $ represents the Gamma distribution. In this way, the correlation between noise distributions among different bands is then rationally encoded. Through introducing a latent variable $ z_{ijk} $, we can equivalently rewrite Eq. (\ref{np}) as the following form:
\begin{equation}\label{e}
\begin{split}
e_{ij} &\sim \prod_{k=1}^K{\mathrm{N}(e_{ij}|\mu_{jk},\tau_{jk}^{-1})}^{z_{ijk}},\\
\bm{z}_{ij} &\sim \mathrm{Multinomial}(\bm{z}_{ij}|\bm{\pi}_j),\\ \bm{\pi}_j &\sim \mathrm{Dir}(\bm{\pi}_j|\alpha_0),
\end{split}
\end{equation}
where $ \mathrm{Multinomial}(.) $ and $ \mathrm{Dir}(.) $ represent the Multinomial and Dirichlet distributions, respectively.  Then, Eq. (\ref{mutao}) and (\ref{e}) together encode the noise structure embedded in a HSI. Fig. \ref{gm} shows the graphical model for noise encoding within the red box. All involved parameters can be inferred from data, as introduced in the following Section II.C.

\subsection{NMoG-LRMF model}
For the prior structure of the underlying clean HSI matrix $ \bm{L} $, we readily employ the low-rank structure to encode it. Specifically, let's consider the following LRMF model:
\begin{equation}
\bm{Y=UV^T+E},
\end{equation}
where $ \bm{L=UV^T} $ implies the low-rank structure underlying the clean HSI.

For most deterministic LRMF model, the rank $ r $ of matrix $ \bm{L} $ is fixed. By modeling the problem into certain generative model~\cite{babacan2012sparse,zhao2014robust}, the rank can also be adaptively learned from data. Specifically, suppose the columns of $ \bm{U} $ and $ \bm{V} $  are generated from Gaussian distribution. For $ l = 1,2,\cdots, R $, where $  R $ is a preset larger value beyond the true rank $ r $, then
\begin{equation}\label{u,v}
\bm{u}_{\cdot l} \sim \mathrm{N}(\bm{u}_{\cdot l}|0,\gamma_{l}^{-1}I_N),~
\bm{v}_{\cdot l} \sim \mathrm{N}(\bm{v}_{\cdot l}|0,\gamma_{l}^{-1}I_B),
\end{equation}
where $ \bm{u}_{\cdot l} $ and $  \bm{v}_{\cdot l} $ are the $ l^{th} $ columns of $ \bm{U} $ and $ \bm{V} $, respectively. $ I_N(I_B) $ denotes the $ N\times N (B\times B) $ identity matrix. $ \gamma_{l} $ is the precision of  $ \bm{u}_{\cdot l} $ and $ \bm{v}_{\cdot l} $ with prior as follows:
\begin{equation}\label{r}
\gamma_l\sim \mathrm{Gam}(\gamma_l|\xi_0,\delta_0).
\end{equation}
Note that each column pair $ \bm{u}_{\cdot l} $ and $  \bm{v}_{\cdot l} $ of $ \bm{U} $, $ \bm{V} $ has the same sparsity profile characterized by the common precision variable $ \gamma_l $. It has been validated that such a modeling could
lead to large precision values of some $\gamma_{l}$s, and hence is capable of automatically conducting low-rank estimate of $\bm{L}$~\cite{babacan2012sparse}.

Combining Eqs. ($ \ref{np} $)-($ \ref{r} $) together, we can construct the full NMoG-LRMF Bayesian model. And the graphical model representation of this model is shown in Fig. \ref{gm}. The goal turns to infer the posterior of all involved variables:
\begin{equation}\label{post}
p(\bm{U,V,\mathcal{Z},\mu,\tau,\pi,\gamma},d|\bm{Y}),
\end{equation}
where $ \mathcal{Z}=\{\bm{z}_{ij}\}_{N\times B} $, $ \bm{\mu} = \{\mu_{jk}\}_{B\times K} $, $ \bm{\tau} =\{\tau_{jk}\}_{B\times K} $, $ \bm{\pi} = (\bm{\pi}_1,\cdots, \bm{\pi}_B) $, $\bm{\gamma} = (\gamma_1,\cdots,\gamma_R) $.

\subsection{Variational Inference}

We use variational Bayes (VB) method~\cite{bishop2006pattern} to infer the posterior. Specifically, VB aims to use a variational distribution  $ q(\bm{\theta})$ to approximate the true posterior $ p(\bm{\theta|D}) $, where $ \bm{\theta} $ denotes the set of parameters and $ \bm{D} $ denotes the observed data. To achieve this goal, we need to solve the following optimization problem:
\begin{equation}\label{Lq}
\min_{q\in \mathbb{C}} KL(q(\bm{\theta})||p(\bm{\theta|D})) = -\int q(\bm{\theta}) \ln\{\frac{p(\bm{\theta|D})}{q(\bm{\theta})} \} d\bm{\theta},
\end{equation}
where $ KL(q||p) $ represents the KL divergence between two distribution $ q$ and $ p $, and $ \mathbb{C} $ denotes the constraint set of probability densities to make the minimization tractable. Assuming $ \mathbb{C} $ is the distribution family which can be factorized with respect to some disjoint groups: $ q(\bm{\theta})=\prod_i q_i(\bm{\theta}_i) $, the closed-form optimal solution $ q_i^*(\bm{\theta}_i) $ can be obtained by
\begin{equation}\label{vi}
q_i^*(\bm{\theta}_i) = \frac{\exp\{ \langle \ln p(\bm{\theta,D}) \rangle _ {\bm{\theta \setminus \theta}_i} \}} {\int {\exp \{ \langle \ln p(\bm{\theta,D}) \rangle _ {\bm{\theta} \setminus \bm{\theta}_i} \}}},
\end{equation}
where $ \langle \cdot \rangle $ denotes the expectation and $ \bm{\theta} \setminus \bm{\theta}_i $ denotes the set of $ \bm{\theta}  $ with $ \bm{\theta}_i $ removed.

Then we can analytically infer all the factorized distributions involved in Eq. (\ref{post}). Suppose that the approximation of posterior distribution ($\ref{post}$) possesses a factorized form as follows:
\begin{equation}\label{est}
\begin{split}
q(\bm{U,V,\mathcal{Z},\mu,\tau,\pi,\gamma,d})=\prod_i{q(\bm{u}_{i \cdot})} &
\prod_j{q(\bm{v}_{j \cdot})} \\
\prod_{ij}{q(\bm{z}_{ij})}   \prod_j{q(\bm{\mu_j,\tau_j)}q(\bm{\pi_j})}  \prod_l&{q(\gamma_l)} q(d)   ,
\end{split}
\end{equation}
where $ \bm{u}_{i \cdot} $ $ (\bm{v}_{j \cdot} )$ are the $i^{th}$ $(j^{th})$ row of matrix $ \bm{U} $ $( \bm{V}) $, respectively. According to Eq. ($ \ref{vi} $), we can get the closed-form inference equation for each component of (\ref{est})\footnote{Inference details to obtain these equations can be referred to in http://dymeng.gr.xjtu.edu.cn.}.

\textbf{Estimation of noise component:} We first list the updating equation for the noise components involved in Eq. (\ref{est}). The posterior distribution of mean and precision for noise in each band is updated by the following equation:
\begin{equation}\label{mu,tau}
q^*(\bm{\mu}_j,\!\bm{\tau}_j)
\!=\! \prod_k\!{\mathrm{N}(\mu_{jk}|m_{jk},\!\frac{1}{\beta_{jk}\tau_{jk}}) \mathrm{Gam}(\tau_{jk}|c_{jk},d_{jk})},
\end{equation}
where the parameters in the above equation can be calculated from data in the following way:
\begin{equation}  \label{mutau1}
m_{jk} = \frac{1}{ \beta_{jk} } \{ \beta_0 m_0 \sum\nolimits_{i}{ \langle z_{ijk} \rangle (Y_{ij}- \langle \mu_{jk}  \rangle )} \},
\end{equation}
\begin{equation}\label{mutau2}
\beta_{jk} = \beta_0+ \sum\nolimits_{i}{ \langle z_{ijk} \rangle },~
c_{jk} = c_0 + \frac{1}{2}\sum\nolimits_{i}{ \langle z_{ijk} \rangle },
\end{equation}
\begin{equation} \label{mutau3}
\begin{split}
d_{jk}  &=  \langle d \rangle + \frac{1}{2} \{ \sum\nolimits_{i}{\langle z_{ijk} \rangle  \langle (Y_{ij}- \bm{u}_{i\cdot}\bm{v}_{j\cdot}^T)^2 \rangle}  +\beta_0m_0^2 \\
&\qquad - \frac{1}{\beta_{jk}} [ \sum\nolimits_{i}{ \langle z_{ijk} \rangle (Y_{ij}- \bm{u}_{i\cdot}\bm{v}_{j\cdot}^T )} + \beta_0^2m_0^2 ]^2 \}.
\end{split}
\end{equation}

The update equation of the latent variable $ \mathcal{Z} $ is:
\begin{equation} \label{z}
q^*(\bm{z}_{ij}) = \prod\nolimits_k{\varrho_{ijk}^{z_{ijk}}},
\end{equation}
where the involved parameters are calculated by
\begin{equation} \label{z1}
\begin{split}
\varrho_{ijk}& = \rho_{ijk}/ \sum\nolimits_k{\rho_{ijk}},\\
\ln\rho_{ijk} & =  \langle \ln \pi_{jk} \rangle - \ln \sqrt{2\pi} + \langle \ln \tau_{jk} \rangle/2 \\
&\qquad- \langle \tau_{jk}(Y_{ij}-\mu_{jk}-\bm{u}_{i\cdot}\bm{v}_{j\cdot}^T)^2 \rangle/2.
\end{split}
\end{equation}

Similarly, the update equation for the mixing proportion $ \bm{\pi}_j $ over the $ j^{th} $ band can be written as:
\begin{equation}\label{pi}
q^*(\bm{\pi}_{j}) = \prod\nolimits_k{\pi_{jk}^{\alpha_{jk}-1}},
\end{equation}
where $\alpha_{jk}=\alpha_{0k}+\sum_i{\langle z_{ijk} \rangle}$.

The update equation on the hyper-parameter  $ d $ is:
\begin{equation}\label{d}
q^*(d) = \mathrm{Gam}(d|\eta,\lambda),
\end{equation}
where $\eta=\eta_0+c_0KB$ and $\lambda=\lambda_0+\sum_{j,k}{\langle \tau_{jk} \rangle}$.

\textbf{Estimation of low-rank component.} Completing the component update for noise, we then estimate the posterior of low-rank component $ \bm{u}_{i \cdot}(i=1,\cdots,N)$ and $ \bm{v}_{j \cdot}(j=1,\cdots,B) $ as:
\begin{equation}\label{u}
q^*(\bm{u}_{i \cdot})=\mathrm{N}(\bm{u}_{i \cdot}|\bm{\mu}_{\bm{u}_{i \cdot}},\bm{\Sigma}_{\bm{u}_{i \cdot}}),
\end{equation}
where
\begin{equation*} \label{u1}
\begin{split}
\bm{\mu}_{\bm{u}_{i \cdot }} &= \{ \sum\nolimits_{j,k}{ \langle z_{ijk} \rangle \langle \tau_{jk} \rangle   (Y_{ij}- \langle \mu_{jk} \rangle)  \langle \bm{v}_{j \cdot } \rangle } \}\bm{\Sigma}_{\bm{u}_{i \cdot}},\\
\bm{\Sigma}_{\bm{u}_{i \cdot}} &=  \{ \sum\nolimits_{j,k}{ \langle z_{ijk} \rangle   \langle \tau_{jk} \rangle   \langle \bm{v}_{j\cdot }^T \bm{v}_{j\cdot } \rangle  +  \langle \bm{\Gamma} \rangle }  \}^{-1}.
\end{split}
\end{equation*}

\begin{equation}\label{v}
q^*(\bm{v}_{j \cdot})= \mathrm{N}(\bm{v}_{j \cdot}|\bm{\mu}_{\bm{v}_{j \cdot}},\bm{\Sigma}_{\bm{v}_{j \cdot}}),
\end{equation}
where
\begin{equation} \label{vste}
\begin{split}
\bm{\mu}_{\bm{v}_{j \cdot }} &= \{ \sum\nolimits_{i,k}{ \langle z_{ijk} \rangle \langle \tau_{jk} \rangle   (Y_{ij}- \langle \mu_{jk} \rangle )  \langle \bm{u}_{i \cdot } \rangle } \}\bm{\Sigma}_{\bm{v}_{j \cdot}},\\
\bm{\Sigma}_{\bm{v}_{j \cdot}} &=  \{ \sum\nolimits_{i,k}{ \langle z_{ijk} \rangle   \langle \tau_{jk} \rangle   \langle \bm{u}_{i\cdot }^T \bm{u}_{i\cdot } \rangle  +  \langle \Gamma \rangle }  \}^{-1},
\end{split}
\end{equation}
where $ \bm{\Gamma} = \mathrm{diag}(\langle \bm{\gamma} \rangle)$. For $ \gamma_l $ which controls the rank of $ \bm{U} $ and $ \bm{V} $, we have :

\begin{equation} \label{gam}
q^*(\gamma_l) = \mathrm{Gam}(\gamma_r|\xi_l,\delta_l),
\end{equation}
where
\begin{equation*} \label{gam1}
\begin{split}
\xi_l &= \xi_0 + (m+n)/2,\\
\delta_l & = \delta_0 + \sum\nolimits_{i}{\langle u_{il}^2 \rangle}/2 +\sum\nolimits_{j}{\langle v_{jl}^2 \rangle}/2.
\end{split}
\end{equation*}
$ m $ and $ n $ represent the image size among the spatial dimension. As discussed by Babacan et al.~\cite{babacan2012sparse}, some $\gamma_l$s tend to
be very large during the inference process and the corresponding  rows will be removed from $ \bm{U} $ and $ \bm{V} $. The low-rank purpose can thus be rationally conducted. In all our experiments, we just automatically infer the rank of the reconstructed matrix through throwing away those comparatively very large $ \gamma_l $ as previous literatures did~\cite{zhao2014robust}.

The proposed variational inference method for the NMoG-LRMF model can then be summarized in Algorithm \ref{alg1}.

\begin{algorithm}
	\small
	\caption{NMoG-LRMF algorithm}\label{alg1}
	\begin{algorithmic}[1]
		\Require  the original HSI matrix $\bm{Y}\in \mathbb{R}^{N\times B}$, Gaussian component number $ K $, and maximum iteration number.
		
		\Ensure $\mathbf{U_{opt}}=\mathbf{U}_{t}$, $\mathbf{V_{opt}}=\mathbf{V}_{t}$.
		\renewcommand{\algorithmicensure}{\textbf{Initialization:}}
		
		\Ensure Parameters $ (m_0,\beta_0,c_0,d_0 ,\eta_0,\lambda_0) $ in noise prior. Low-rank components $\mathbf{U}_0,\mathbf{V}_0$ and parameters in model prior $ (\xi_0,\delta_0) $; $ t=1 $.
		
		\While { not coverged }
		\State Update approximate posterior of noise component $ \mathcal{Z}^t, \bm{\pi^t} $ by Eq. ($ \ref{z} $)-($ \ref{pi} $).		
		\State Update approximate posterior of noise component $ \bm{\mu}^t, \bm{\tau^t} $ by Eq. ($ \ref{mu,tau} $)-($ \ref{mutau3}) $.		
		\State Update approximate posterior of noise component $ d^t $ by Eq. ($ \ref{d} $).		
		\State Update approximate posterior of low-rank \mbox{component} $ \bm{U}^t, \bm{V^t} $ by Eq. ($ \ref{u} $)-(\ref{vste}).			
		\State Update approximate posterior of parameters in noise component $ \bm{\gamma}^t $ by Eq. ($ \ref{gam}$).						
		\small
		\State $t\leftarrow t+1$.
		\EndWhile
	\end{algorithmic}
	\normalsize
\end{algorithm}

\textbf{Setting of hyper-parameters: } We set all the hyper-parameters
involved in our model in a non-informative manner to make them possibly less affect the inference of posterior distributions \cite{bishop2006pattern}. Throughout our experiments, we set $ m_0 $ as 0, and $ \beta_0, c_0, d_0 ,\eta_0, \lambda_0, \xi_0, \delta_0$ as a small value $ 10^{-3} $. Our method performs stably well under such easy settings.

\section{Experimental Results}

In this section, to evaluate the performance of the proposed NMoG-LRMF method, we conducted a series of experiments on both synthetic and real HSI data. Compared methods include  LRMR~\cite{zhang2014hyperspectral} and LRTV~\cite{hetotal}, considering  deterministic Gaussian noise and sparse noise.  Meanwhile, five representative low-rank matrix analysis methods considering different kinds of i.i.d. noise distributions were also considered for comparison, including  PMoEP~\cite{cao2015PMoEP} (assuming i.i.d. mixture of Exponential Power noise), MoG-RPCA ~\cite{zhao2014robust} (assuming i.i.d. MoG noise), RegL1ALM~\cite{zheng2012practical}, CWM~\cite{meng2013cyclic} (assuming i.i.d. Laplace noise) and SVD~\cite{haardt1996method} (assuming i.i.d. Gaussian noise).  Besides, the performance of TDL~\cite{peng2014decomposable} and BM4D~\cite{BM4D2013TIP} are also compared, and both methods represent the state-of-the-art methods for HSI denoising by considering HSI priors. All experiments were implemented in Matlab R2014b on a PC with 4.0GHz CPU and 31.4GB RAM.

\subsection{Simulated HSI denoising experiments}

In this experiment, we focus on the performance of NMoG-LRMR in HSI denoising  with syntheic noise. Two HSIs were employed: Washington DC Mall\footnote{\url{http://engineering.purdue.edu/~biehl/MultiSpec/hyperspectral.html}} with size of $ 1208 \times 307 \times 191$ and RemoteImage\footnote{\url{http://peterwonka.net/Publications/code/LRTC\_Package\_Ji.zip}} provided by Liu et. al. \cite{liu2013tensor} with size of $ 205 \times 246 \times 96 $.  After cropping the main part of HSI and deleting some evident visual contaminative spectral channels, Washington DC Mall and RemoteImage are  resized to $ 200 \times 200 \times 160 $ and $ 200 \times 200 \times 89 $, respectively. The gray value of each band are normalized into $ [0,1] $.

\begin{table*}
	\caption{\label{table2} Performance comparison of all competing methods on DCmall data with synthetic noise. }
	\setlength{\tabcolsep}{4pt}
	\begin{center}
		{\small
			\scalebox{0.9}[0.8]{			
				\begin{tabular}{c c c c c c c c c c c c}
						
						\toprule
						& Noisy HSI  & ~SVD~  & RegL1ALM & CWM & MoG-RPCA	& PMoEP & LRMR & LRTV &TDL & BM4D & NMoG\\
						\midrule
						\multicolumn{11}{c}{i.i.d. Gaussian Noise}\\ \hline
						
						MPSNR & 26.02  & 44.76  & 41.19 & 42.89   & 45.26 & 44.26 & 42.48   & 43.83 & \bf{46.08}  & 39.33 & 45.27\\
						MSSIM &  0.513  & 0.983 & 0.965  & 0.973 & 0.985 &  0.980 & 0.972  & 0.977   & {\bf{0.989}} &0.948& 0.985 \\
						time & 	- & \bf{0.290}  & 221.2 & 851.1	& 59.1	& 201.7	& 5367.0 & 512.3 & 44.9	& 965.2   & 77.8  \\  \hline
						
						\multicolumn{11}{c}{Non i.i.d. Gaussian Noise}\\ \hline						
						MPSNR & 40.96  & 43.34  & 47.15 & 45.00  & 46.15 & 44.09 & 48.26  & 47.19 & 41.80  & 45.50 & {\bf{51.11}}\\
						MSSIM & 0.890  & 0.973  & 0.985 & 0.980 & 0.981 &  0.973 &0.987 & 0.986 & 0.903 &0.983 & {\bf{0.990}} \\
						time & 	- & \bf{0.292} 	& 212.0 &800.2	& 121.5	& 1672.3	& 778.6	& 478.9 & 821.2	& 940.8  & 106.7 \\
						
						\hline
						\multicolumn{11}{c}{Gaussian + Stripe Noise}\\ \hline
						MPSNR &44.72  & 46.14   & 48.89 & 48.74 & 48.54 & 48.08 & 51.11  & 49.06 & 44.94   & 46.66 & {\bf{54.16}}\\
						MSSIM & 0.941   & 0.989 & 0.992 & 0.992 & 0.992  &  0.990 & 0.994 &  0.993 & 0.942   & 0.985 & {\bf{ 0.997}} \\
						time & 	- & \bf{0.515}	& 238.5 & 782.6	& 307.6 & 1505.5	& 414.4	& 683.2 & 843.1 & 369.5  & 334.7 \\	
						\hline
						\multicolumn{11}{c}{Gaussian + Deadline noise}\\ \hline
						MPSNR & 44.36  & 44.08    & 48.95 & 47.81 & 48.65 & 46.95  &  50.34  & 48.95 & 44.84 & 46.02 & {\bf{54.04}}\\
						MSSIM & 0.938  & 0.980    & 0.993 & 0.990    & 0.991 & 0.986  &  0.991 & 0.993 & 0.943 & 0.979 & {\bf{0.997}} \\
						time & 	- & \bf{0.291 } &177.6   &690.4	& 136.2	& 2004.7 	&670.8  	& 321.2  &  772.3	&307.5    &143.6   \\					\hline
						\multicolumn{11}{c}{Gaussian + Impluse Noise}\\ \hline
						MPSNR & 43.05 & 44.05  & 49.16 & 47.51 & 49.89 & 46.79  & 51.05  & 49.11 & 43.65 & 45.77 & {\bf{ 52.96}}\\
						MSSIM & 0.925   & 0.981 & 0.993 & 0.991 &0.993  & 0.986 & 0.994  &0.993  & 0.932  & 0.977 & {\bf{0.997}} \\
						time & 	- & \bf{0.293} 	&181.3 &567.4	& 146.5& 1640.9	& 743.6	&501.3 & 695.3	& 1028.6  & 150.2 \\					
						\hline
						\multicolumn{11}{c}{Mixture Noise}\\ \hline
						MPSNR & 42.93 & 43.32 & 48.75   & 44.82 & 47.94   & 44.26 & 48.77  & 48.54 & 43.22 & 43.97   & {\bf{50.38}}\\
						MSSIM & 0.913  &0.977  & 0.991& 0.982 & 0.989 & 0.978  & 0.986 & 0.991 & 0.917   & 0.968 & {\bf{0.996}} \\
						time & 	- & \bf{0.289} 	& 176.5 &688.7& 136.3 	& 940.3	& 708.2 & 358.9	& 653.8  & 325.4&142.0\\				
						\hline									
					\end{tabular} }
				}
			\end{center}
			
		\end{table*}
		
		\begin{table*}
			\caption{\label{table3} Performance comparison of all competing methods on RemoteImage data with synthetic noise.}
			\setlength{\tabcolsep}{4pt}
			\begin{center}
				{\small
					\scalebox{0.9}[0.8]{
						\begin{tabular}{c c c c c c c c c c c c}
								\toprule
								& Noisy HSI  & ~SVD~  & RegL1ALM & CWM & MoG-RPCA	& PMoEP & LRMR & LRTV &TDL & BM4D & NMoG\\
								\midrule
								\multicolumn{11}{c}{i.i.d. Gaussian Noise}\\ \hline
								MPSNR & 26.02   & 38.19  & 36.63 & 36.78    & 38.66 & 36.72  & 37.46    & 37.89 & \bf{39.81}   & 38.98 & 38.65\\
								MSSIM &  0.591    & 0.956 & 0.937    & 0.939 & 0.964  &  0.956 & 0.946  & 0.949   & {\bf{0.971}} &0.971 & 0.964 \\
								time & 	- & \bf{0.247}  	& 105.9 & 331.6	& 65.7	& 38.1 	& 409.1 & 211.9 & 47.7 	& 357.7    & 69.6  \\ \hline
								\multicolumn{11}{c}{Non i.i.d. Gaussian Noise}\\ \hline
								MPSNR & 25.17  & 34.36   & 35.51 & 35.37    & 36.41 & 35.67  &  35.95  & 35.73 & 29.74  & 36.82 & {\bf{38.26}}\\
								MSSIM & 0.539   & 0.902 & 0.921  & 0.922 & 0.942  & 0.928 & 0.926 & 0.927 & 0.729 &0.952 & {\bf{0.962}} \\
								time & 	- & \bf{0.253} 	& 105.4 &329.3	&100.2	& 301.9 & 398.7 &220.2 & 151.0 	& 327.8  & 101.1	
								\\
								
								\hline
								\multicolumn{11}{c}{Gaussian + Stripe Noise}\\ \hline
								MPSNR & 29.08  & 37.48  & 39.31 & 38.64  & 39.70 & 38.62  & 40.13  & 39.36 & 29.76 & 37.03  & {\bf{40.68}}\\
								MSSIM & 0.710  & 0.955    & 0.968 & 0.964    & 0.974 & 0.965  & 0.971 & 0.968 & 0.731   & 0.938 & {\bf{0.983}} \\
								time & 	- & \bf{0.279} 	& 108.9 &375.8	&88.6	& 299.8	&606.4	& 238.3 & 1969.9	& 160.4 &112.3					\\		
								\hline
								\multicolumn{11}{c}{ Gaussian + Deadline noise}\\ \hline
								MPSNR & 27.97  & 33.37  & 39.08 & 37.94   & 39.58 & 38.20  &  38.74  & 38.72 & 29.98   & 34.02 & {\bf{40.43}}\\
								MSSIM & 0.693  & 0.907   & 0.967 & 0.961    & 0.972 & 0.963 & 0.958 & 0.965 & 0.772  & 0.888 & {\bf{0.983}} \\
								time & 	- & \bf{0.299}	& 108.9 & 361.7	&119.6	&  1205.9	& 725.1 &378.9  & 273.5 	& 683.6    & 127.5
								\\
								\hline
								\multicolumn{11}{c}{Gaussian + Impluse Noise}\\ \hline
								MPSNR & 28.20  & 35.00 & 41.89 & 39.48  & 41.15 & 39.15&41.32  & 40.63  & 30.63 & 36.97   & {\bf{42.90}}\\
								MSSIM & 0.660  & 0.917 & 0.983 & 0.971  & 0.981 & 0.969  &  0.977 & 0.977 & 0.745  & 0.951 & {\bf{0.990}} \\
								time & 	- & \bf{0.236} 	& 100.0 & 317.2 &110.1	& 202.8 & 473.1	&251.6 & 190.7	& 445.3 & 118.1
								\\						
								\hline
								\multicolumn{11}{c}{Mixture Noise}\\ \hline
								MPSNR & 25.84   & 31.55  & 37.93 & 35.72   & 38.68 & 36.09 &  37.76  & 38.31 & 26.93   & 30.29 & {\bf{39.32}}\\
								MSSIM & 0.590  & 0.868   & 0.958& 0.936  & 0.966 &  0.940  &  0.944 & 0.962 & 0.633  & 0.782 & {\bf{0.979}} \\
								time & 	- & \bf{0.283} 	& 122.0 &	416.4	& 69.5	& 650.7	&696.4 &260.8 & 312.2 & 86.7 &105.0
								\\						
								\hline						
							\end{tabular}}
						}
					\end{center}					
				\end{table*}
					
Real-world HSIs are usually contaminated by several different types of noise, including the most common Gaussian noise, impulse noise, dead pixels or lines, and stripes \cite{zhang2014hyperspectral}. In order to simulate these real HSI noise scenarios, we added six kinds of noises to the original HSI data.

(1) \textit{I.i.d. Gaussian noise}: Entries in all bands were corrupted by zero-mean i.i.d. Gaussian noise $ \mathrm{N}(0,\sigma^2) $ with $ \sigma = 0.05 $.

(2) \textit{Non-i.i.d. Gaussian noise}: Entries in all bands were corrupted by zero-mean Gaussian noise with different intensity. The signal noise ratio (SNR) value of each band is generated from uniform distribution with value in the range of  $ [5,10]\mathrm{dB} $.

(3) \textit{Gaussian + Stripe noise}: All bands were corrupted by Gaussian noise as Case (2). Besides, 40 bands in DCmall data (30 bands in RemoteImage data) were randomly chosen to add stripe noise. The number of stripes in each band is from 20 to 40.

(4) \textit{Gaussian + Deadline noise}: Each band was contaminated by Gaussian noise as Case (2). 40 bands in DCmall data (30 bands in RemoteImage data) were chosen randomly to add deadline noise. The number of deadline is from 5 to 15.

(5) \textit{Gaussian + Impluse noise}: All bands were corrupted by Gaussian noise as Case (2). 40 bands in DCmall (30 bands in RemoteImage data) were randomly chosen to added impluse noise with different intensity, and the percentage of impluse is from $ 50 \% $ to $ 70 \% $.

(6) \textit{Mixture noise}: Each band was randomly corrupted by at least one kind of  noise mentioned in case (2)-(5).
		
Three criteria were utilized to measure performance: (1) MPSNR \cite{huynh2008scope}: Mean of peak signal-to-noise ratio (PSNR) over all bands between clean HSI and recovered HSI. (2) MSSIM \cite{wang2004image}: Mean of structural similarity (SSIM) between clean HSI and recovered HSI over all bands. (3) Time: Time cost of each method  used to complete the denoising process.

\begin{figure*}[t]
	\centering
	\includegraphics[width=0.9\linewidth]{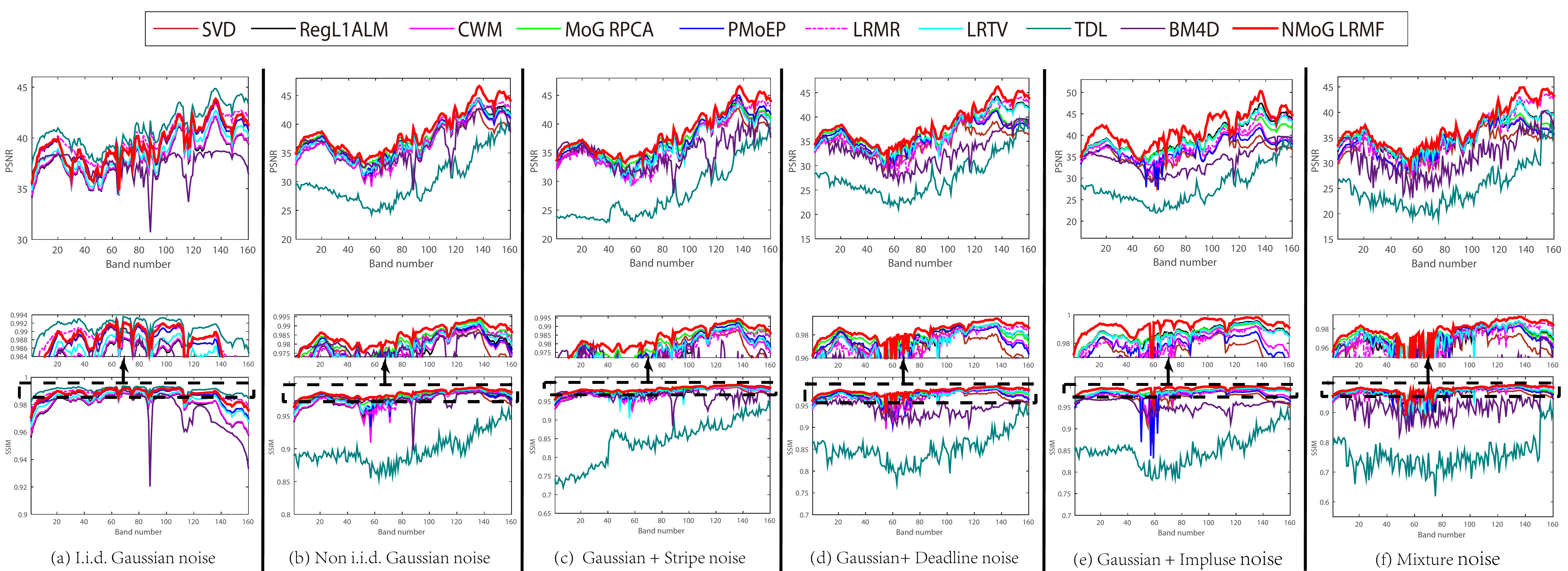}\vspace{-3mm}
	\caption{Each column shows the average PSNR and SSIM measurements among  20 initializations of all methods under certain type of noise in DCmall data: (a) Gaussian noise. (b) Gaussian + Stripe noise. (c) Gaussian + Deadline noise. (d) Gaussian + Impluse noise. (e) Mixture noise. The demarcated area of the subfigure indicates the curve locality on a larger scale. \label{DC_PSNR} }
\end{figure*}

\begin{figure*}[t]
	\centering
	\includegraphics[width=0.9\linewidth]{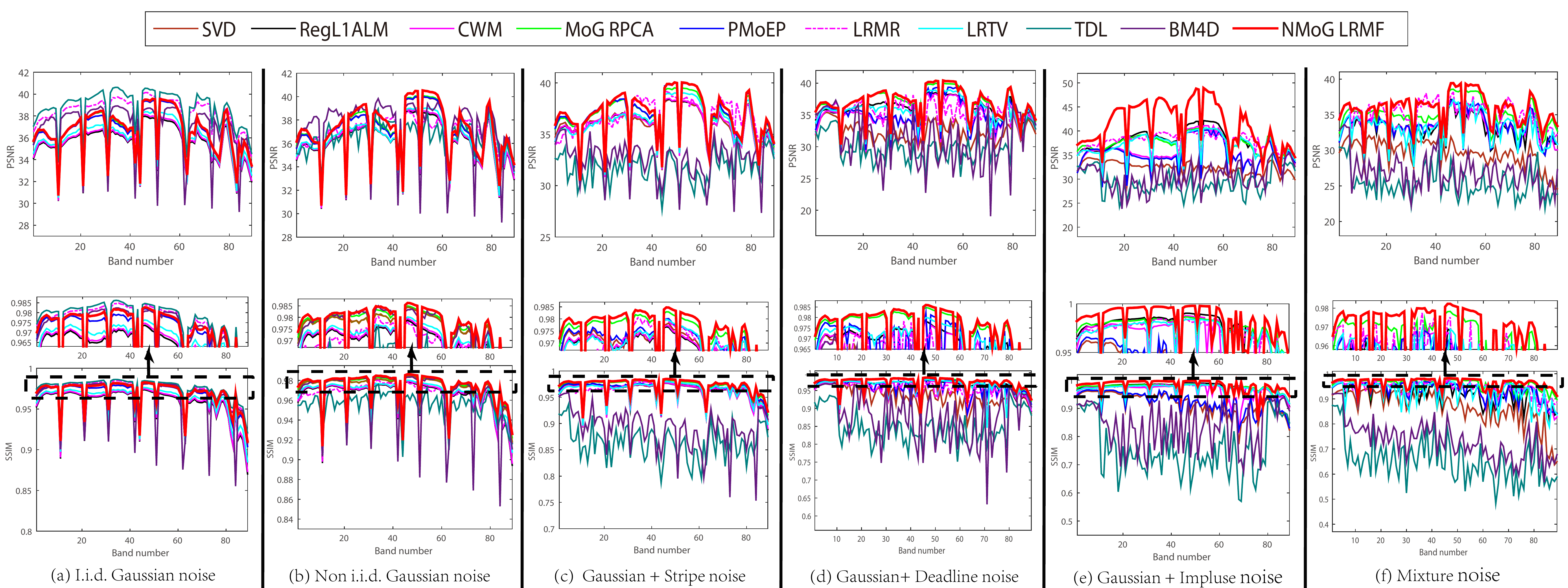}\vspace{-3mm}
	\caption{Each column shows the average PSNR and SSIM measurements among  20 initializations  of all methods under certain type of noise in RemoteImage data: (a) Gaussian noise. (b) Gaussian + Stripe noise. (c) Gaussian + Deadline noise. (d) Gaussian + Impluse noise. (e) Mixture noise.  \label{RM_PSNR} }
\end{figure*}

\begin{figure*}[]
	\centering
	\includegraphics[width=0.8\linewidth]{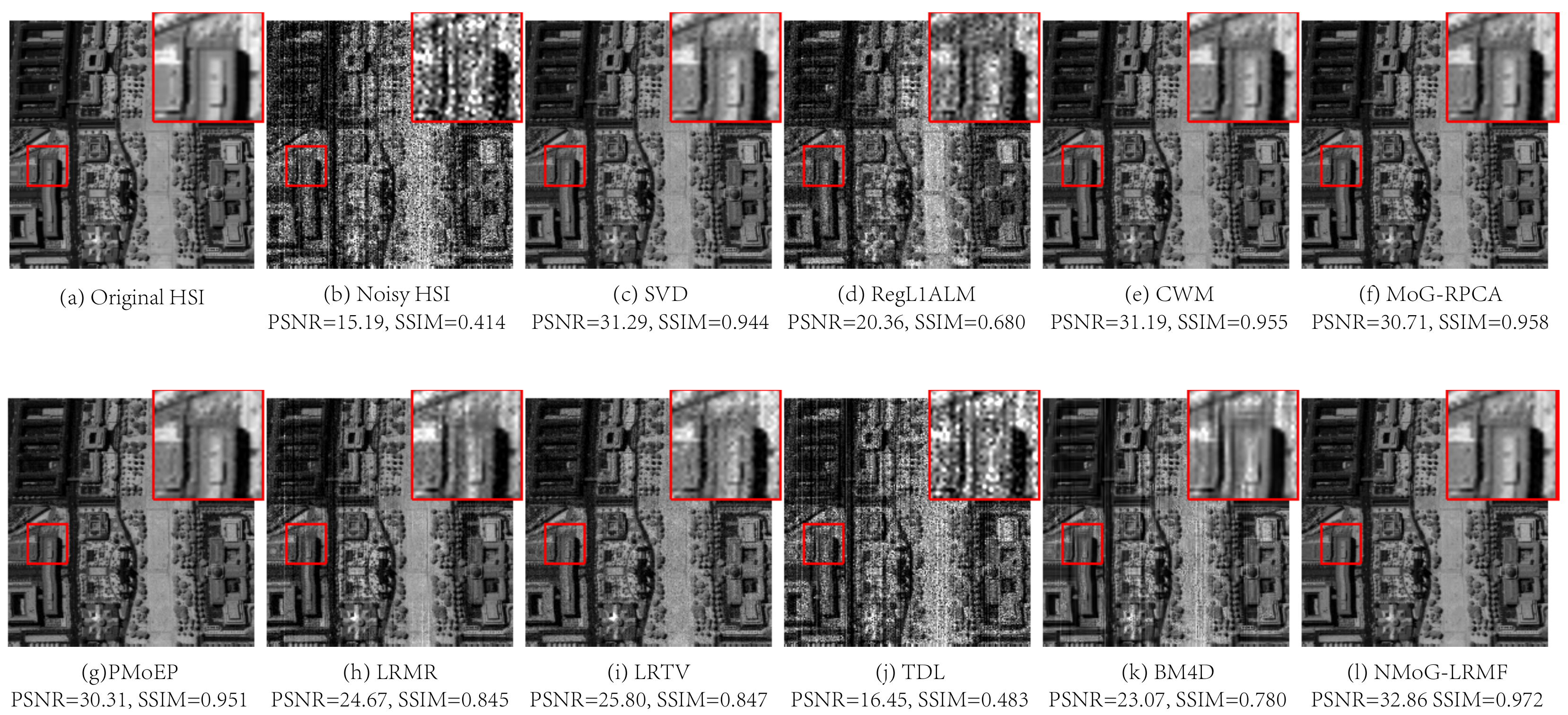}\vspace{-3mm}
	\caption{Restoration results of band 75  under mixture noise in DCmall  data:
		(a) Original HSI, (b) Noisy HSI,  (c) SVD, (d) RegL1ALM, (e) CWM, (f) MoG-RPCA,  (g) PMoEP, (h) LRMR, (i) LRTV, (j) TDL, (k) BM4D, (l) NMoG.
		This figure should be better seen by zooming on a computer screen.\label{DCb72} }
\end{figure*}

\begin{figure*}[]
	\centering
	\includegraphics[width=0.8\linewidth]{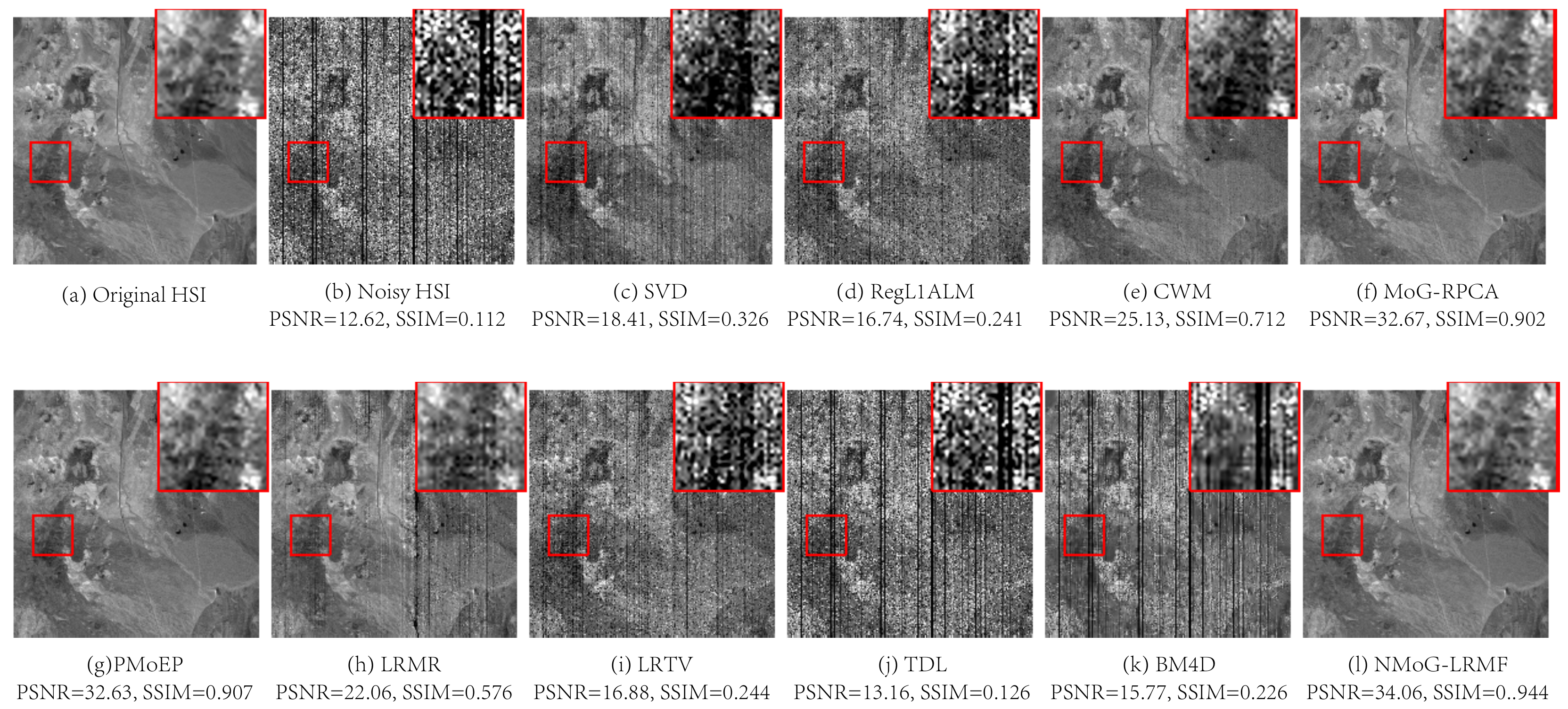}\vspace{-3mm}
	\caption{Restoration results of band 86 under mixture noise in RemoteImage data:
		(a) Original HSI, (b) Noisy HSI,  (c) SVD, (d) RegL1ALM, (e) CWM, (f) MoG-RPCA,  (g) PMoEP, (h) LRMR, (i) LRTV, (j) TDL, (k) BM4D, (l) NMoG.\label{RMb60} }
\end{figure*}

\begin{figure*}
	\centering
	\includegraphics[width=0.8\linewidth]{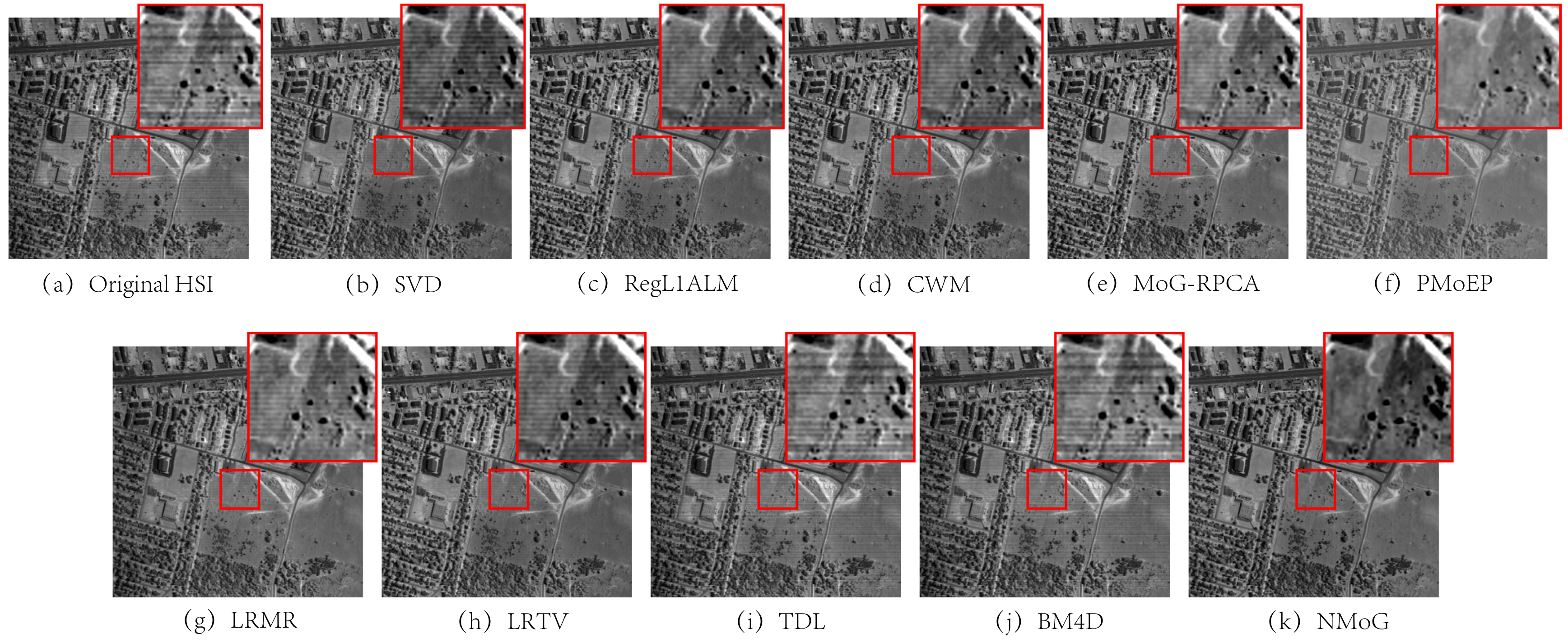}\vspace{-3mm}
	\caption{Restoration results of all methods on band 103 in Urban HSI data. (a) Original HSI, (b) SVD, (c) RegL1ALM, (d) CWM, (e) MoG-RPCA, (f) PMoEP, (g) LRMR, (h) LRTV, (i) TDL, (j) BM4D, (k) NMoG.  \label{urban103} }
\end{figure*}

\begin{figure*}
	\centering
	\includegraphics[width=0.8\linewidth]{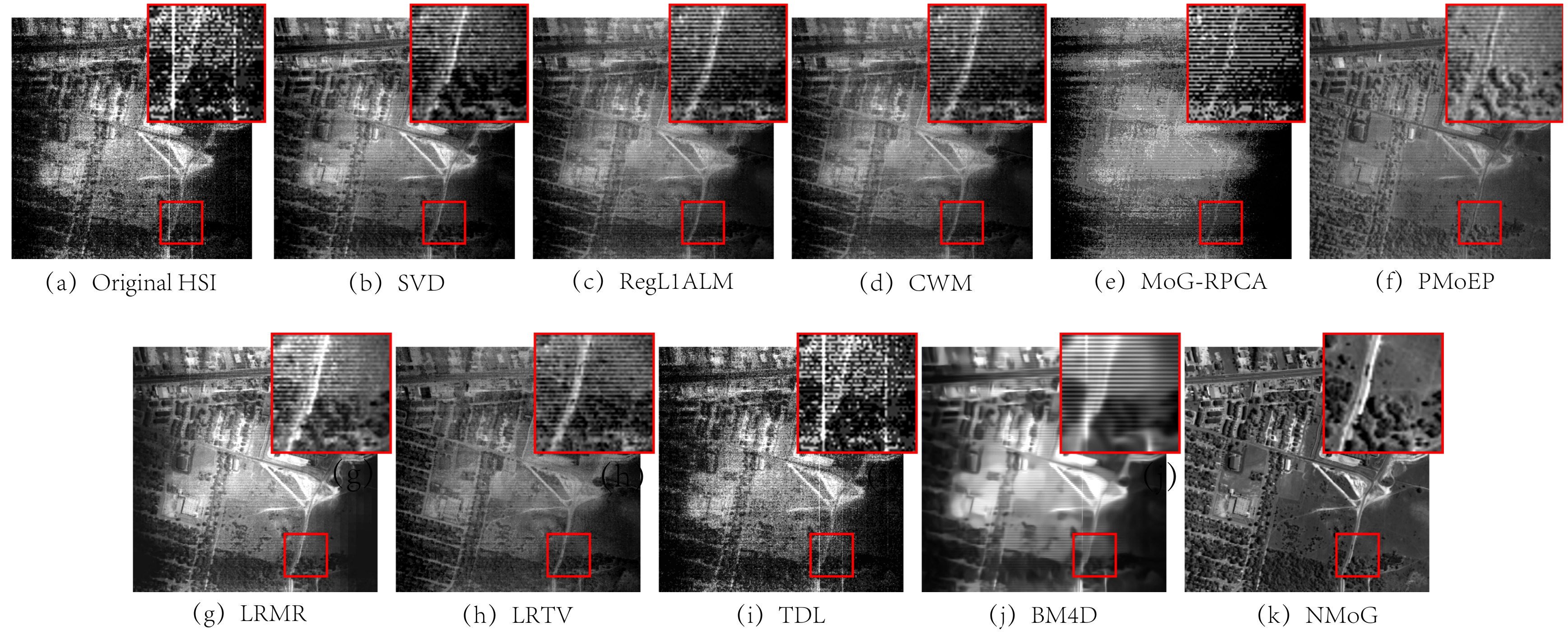}\vspace{-3mm}
	\caption{Restoration results of all methods on band 139 in Urban HSI data. (a) Original HSI, (b) SVD, (c) RegL1ALM, (d) CWM, (e) MoG-RPCA, (f) PMoEP, (g) LRMR, (h) LRTV, (i) TDL, (j) BM4D, (k) NMoG.  \label{urban139} }
\end{figure*}

The parameters of competing methods are set as follows:  the block size in LRMR is $ 20 \times 20 \times B $ and the step size is 10. For LRTV, $\epsilon_1=\epsilon_2=10^{-8} $ and $ \lambda=1/\sqrt{NB} $. For TDLd, we set the block size as $ 6 \times 6 \times B $ and the step size is 2. For BM4D, we set the block size as $ 8 \times 8 \times 8 $ and the step size is 4.  For NMoG-LRMF method, The component number $ K $ in each band was fixed as 1 in Case (1)-(2), and 3 in Case (3)-(6). 
The rank of all low-rank based methods is set as 5 in DCmall data experiment and 4 in RemoteImage data experiment.
All parameters involved in the competing methods were empirically tuned or specified as suggested by the related literatures to guarantee their possibly good performance.
All competing methods run with 20 random initializations in each noise case, and the average result is reported.

The results of all competing methods in DCmall and RemoteImage HSI data are shown in Table \ref{table2} and Table \ref{table3}, respectively.	
The superiority of the proposed method can be easily observed, except in the i.i.d. Gaussian noise case, which complies with the basic noise assumption of conventional methods. In i.i.d. Gaussian case, instead of only using the simple low-rank prior in our method, multiple competing methods, like TDL and BM4D, utilize more useful HSI priors in their model, and thus tend to have relatively better performance. While on more complex but more practical complicated non-i.i.d. noise cases, the advantage of the proposed method is evident. This can be easily explained by the better noise fitting capability of the proposed method, i.e., it can more properly extract noises embedded in HSI, which then naturally leads to its better HSI recovery performance.

Furthermore, it also can be seen that the computational cost of the proposed method is with almost similar order of magnitude with other competing methods, except the known SVD, which we use the mature toolkit in Matlab and can be implemented very efficiently. Considering its better capability in fitting much wider range of noises than current methods, it should be rational to say that the proposed method is efficient.

We further show the PSNR and SSIM measurements across all bands of the HSI under six types of noise settings in two experiments in Fig. \ref{DC_PSNR} and Fig. \ref{RM_PSNR}, respectively. From the figures,  it is easy to see that TDL obtains  the best PSNR and SSIM values across all bands in the i.i.d. Gaussian noise.
In other more complex noise cases, NMoG-LRMF achieves the best PSNR and SSIM values across almost all bands. This verifies the robustness of the proposed method over entire HSI bands.

Figs. \ref{DCb72} and \ref{RMb60} give the restoration results of two typical bands in DCmall and RemoteImage HSIs, respectively. The original HSIs are corrupted by mixture of Gaussian, stripes, deadline and impluse noise. It can be observed that the competing methods SVD, CWM, PMoEP, LRMR, and BM4D can hardly remove this complex mixture noise from the images. We can also see that although RegL1ALM, MoG-RPCA and LRTV have a better denoising performance, the restored HSI still relatively blur many details as compared with the ground truth, which are also reflected from their PSNR and SSIM values in the band.
Comparatively, our NMoG-LRMF method achieves better reconstruction effect both visually and quantitatively, and both finely removes the unexpected noises, and better recovers HSI details like edges and textures.

\subsection{Real HSI denoising experiments}
In this section, we evaluate the performance of the proposed method on two real HSI datasets, Urban dataset\footnote{\url{http://www.tec.army.mil/hypercube} } and EO-1 Hyperion dataset\footnote{\url{http://datamirror.csdb.cn/admin/dataEO1Main.jsp}}.

\begin{figure*}
	\centering
	\includegraphics[width=0.8\linewidth]{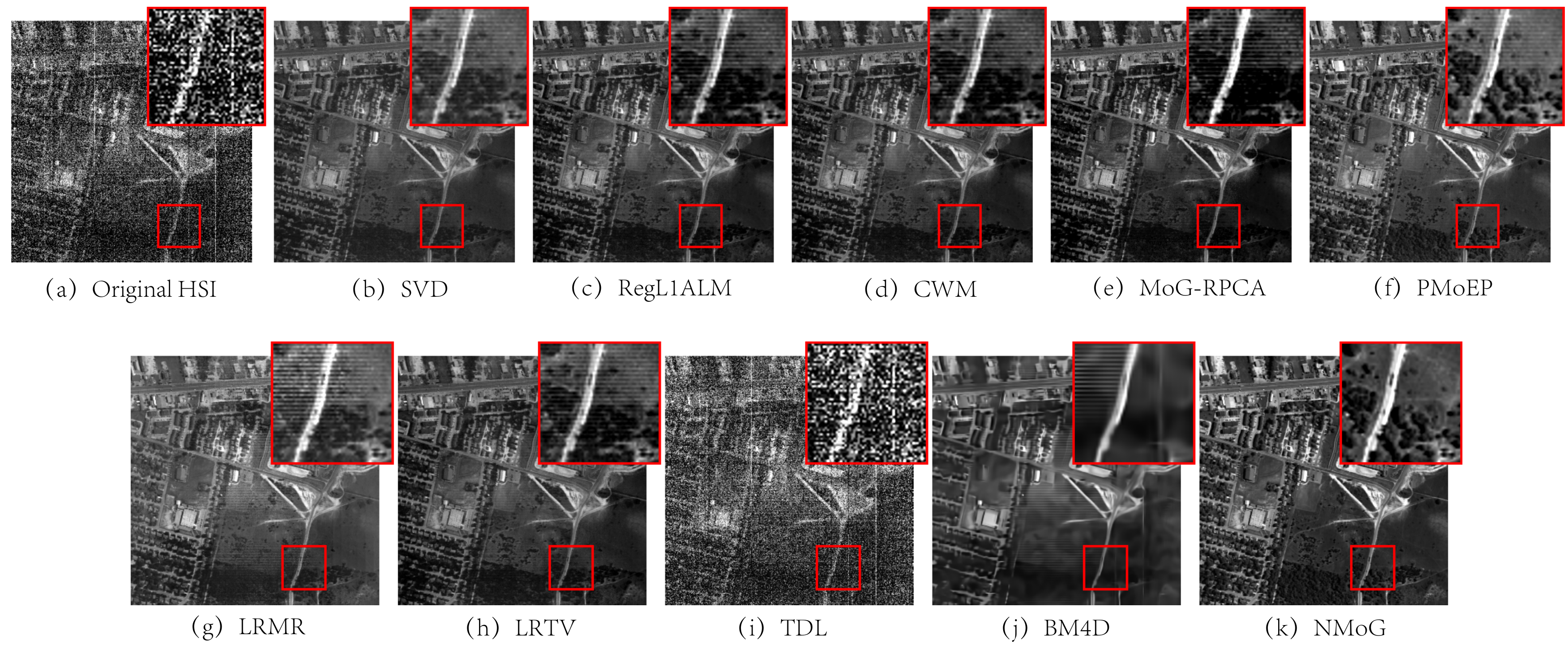}\vspace{-3mm}
	\caption{Restoration results of all methods on band 207 in Urban HSI data.(a) Original HSI, (b) SVD, (c) RegL1ALM, (d) CWM, (e) MoG-RPCA,  (f) PMoEP, (g) LRMR, (h) LRTV, (i) TDL, (j) BM4D, (k) NMoG.  \label{urban207} }
\end{figure*}

\begin{figure*}
	\centering
	\includegraphics[width=0.8\linewidth]{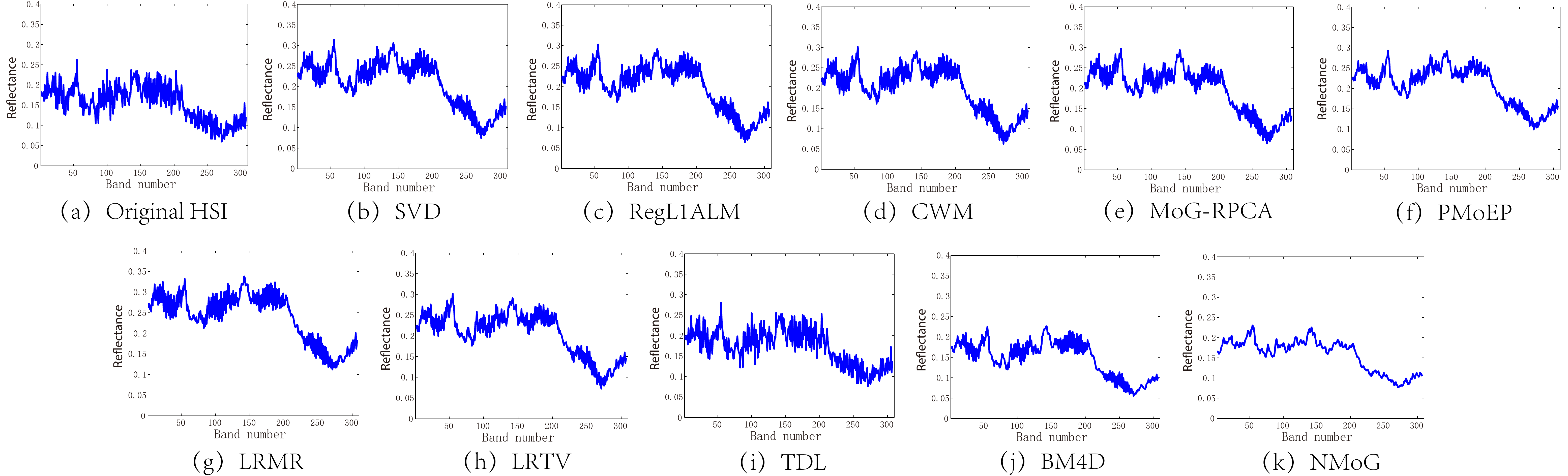}\vspace{-3mm}
	\caption{Horizontal mean profiles of band 207 in Urban HSI data. (a) Original HSI, (b) SVD, (c) RegL1ALM, (d) CWM, (e) MoG-RPCA,  (f) PMoEP, (g) LRMR, (h) LRTV, (i) TDL, (j) BM4D, (k) NMoG. \label{urbanhor} }
\end{figure*}

The similar competing methods as the last section were compared in the real HSI experiments. Like the synthetic experiments, all involved parameters have been finely tuned or directly suggested by the original references to promise a possibly well performance of all competing methods. The component number of each band in the proposed method is easily set as $ 3 $ throughout all experiments. The gray value of HSI were normalized into $ [0,1] $.

The first Urban dataset is of the size $ 307 \times  307 \times  210 $,  and some bands are seriously polluted by atmosphere and water. We use all of data without removing any bands to more rationally verify the robustness of the proposed method in the presence of such heavy noises.

Figs. \ref{urban103}, \ref{urban139} and \ref{urban207} present bands 103, 139 and 207 of the restored images obtained by all competing methods, respectively. From Fig. \ref{urban103}-\ref{urban207} (a), we can see that the original HSI bands are  contaminated by complex structural noise, like the stripe noise. It can be easily observed from Fig. \ref{urban103}-\ref{urban207} (b)-(f)  that the LRMF methods with different i.i.d. noise distribution assumptions cannot finely restore a clean HSI. This can be easily explained by the fact that such structural noises is obvious non-i.i.d. and the deviation between the real noise configuration and the encoded knowledge in the model then naturally degenerate the performance of the corresponding methods. Comparatively, albeit considering less prior knowledge on the to-be-recovered HSI, our method can still achieve a better recovery effect in visualization in its better restoration of texture and edge details and less preservation of structural noises due to its powerful noise modeling ability. This further substantiates the robustness of the proposed method in practical scenarios.

Then we give some quantitative comparison by showing the horizontal mean profiles of band 207 in Urban dataset before and after restoration in Fig. \ref{urbanhor}. The horizontal axis in the figure represents the row number, and the vertical axis represents the mean digital number value of each row. As shown in Fig. \ref{urbanhor}(a), due to the existence of mixed noise, there are rapid fluctuations in the curve. After the restoration processing, the fluctuations are more or less suppressed. It is easy to observe that the restoration curve of our NMoG-LRMF method provides evidently smoother curves.

\begin{figure*}
	\centering
	\includegraphics[width=0.8\linewidth]{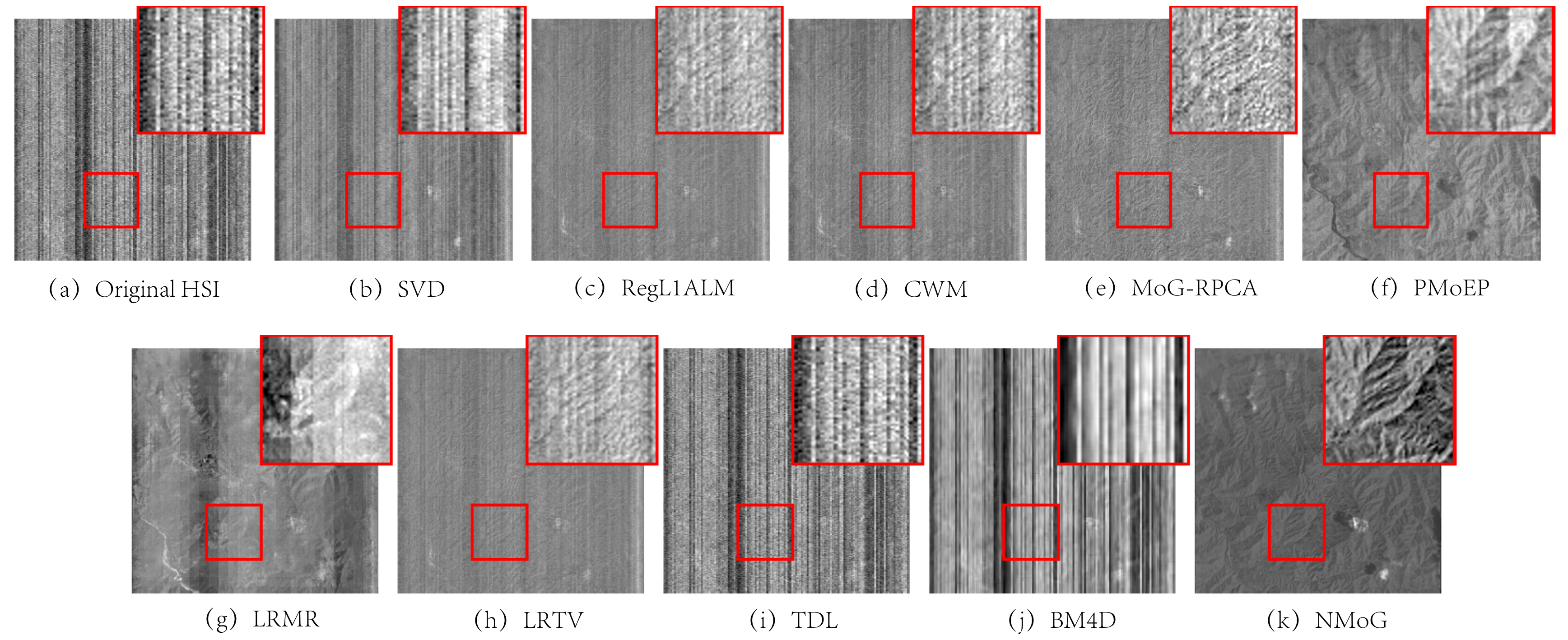}\vspace{-3mm}
	\caption{Restoration results of all methods on band 100 in EO-1 Hyperion data.  (a) Original HSI, (b) SVD, (c) RegL1ALM, (d) CWM, (e) MoG-RPCA, \hspace{1cm} (f) PMoEP, (g) LRMR, (h) LRTV, (i) TDL, (j) BM4D, (k) NMoG. \label{EO1100} }
\end{figure*}

\begin{figure*}
	\centering
	\includegraphics[width=0.8\linewidth]{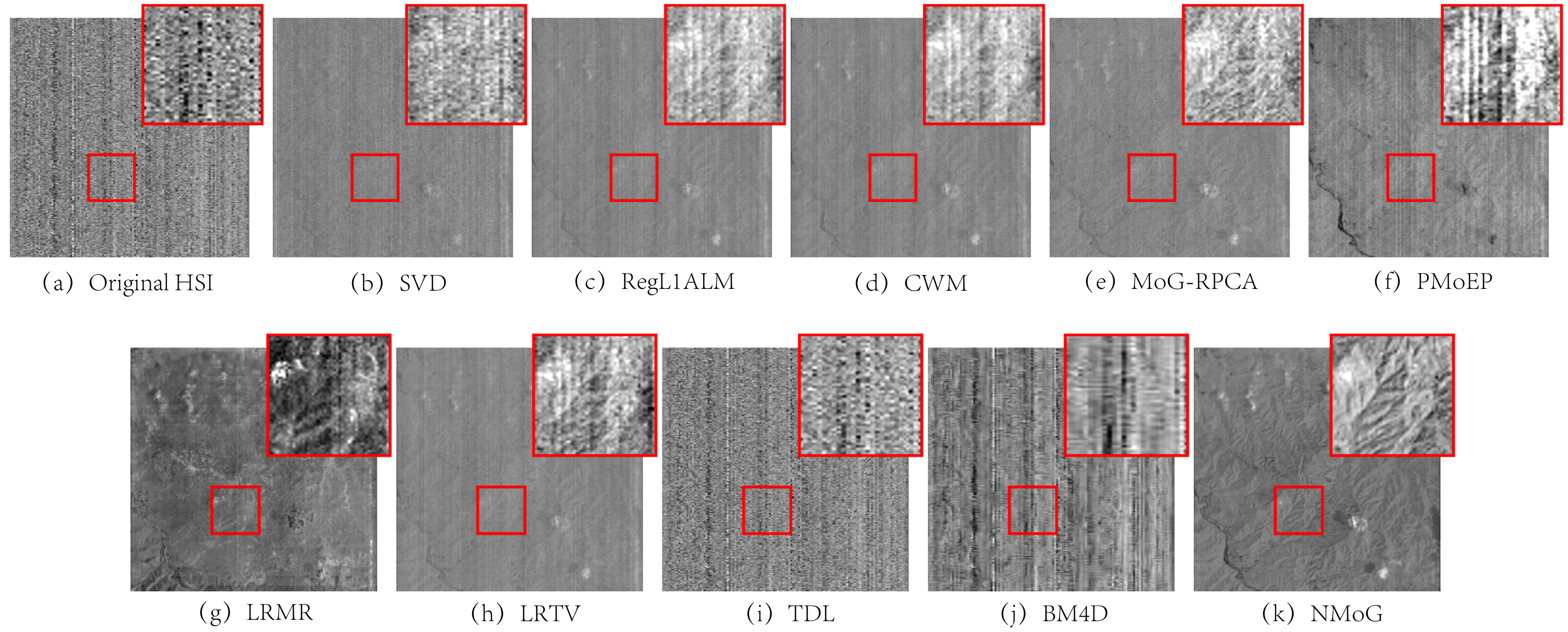}\vspace{-3mm}
	\caption{Restoration results of all methods on band 144 in EO-1 Hyperion data.  (a) Original HSI, (b) SVD, (c) RegL1ALM, (d) CWM, (e) MoG-RPCA, \hspace{1cm} (f) PMoEP, (g) LRMR, (h) LRTV, (i) TDL, (j) BM4D, (k) NMoG.\label{EO1144} }
\end{figure*}

\begin{figure*}
	\centering
	\includegraphics[width=0.8\linewidth]{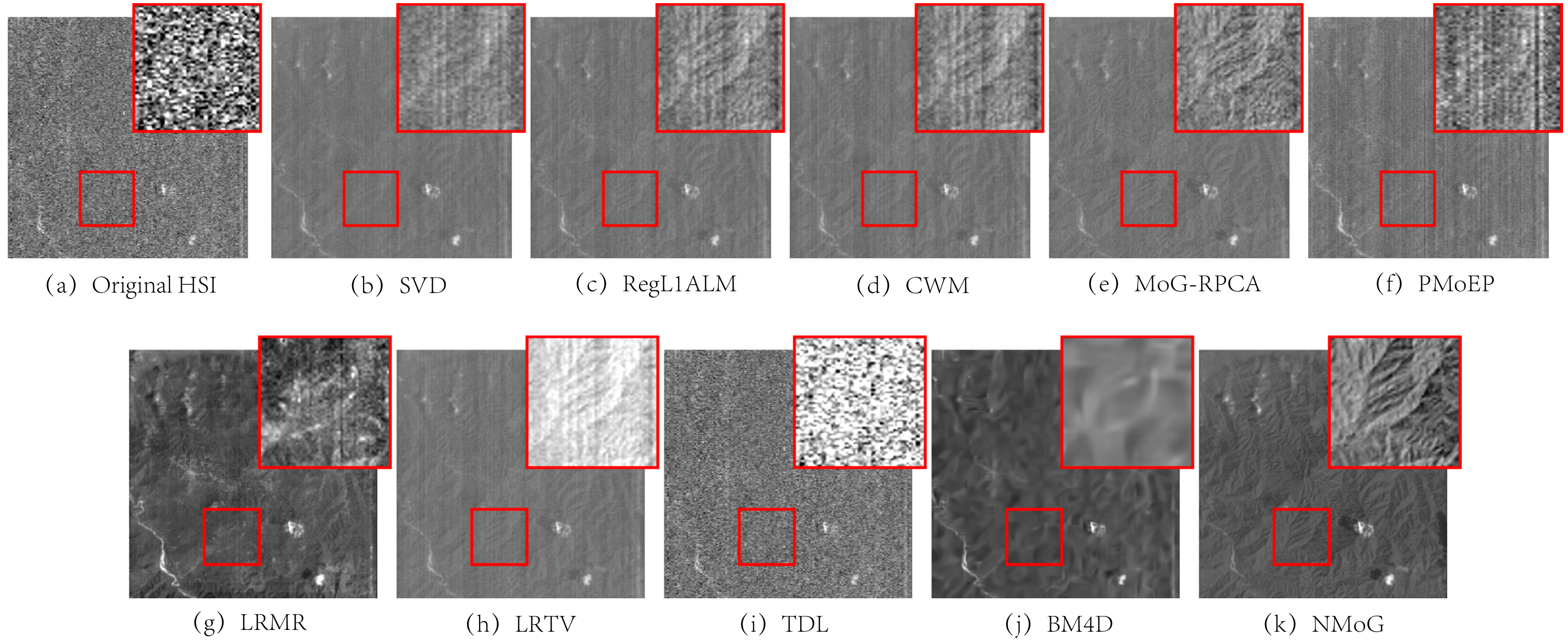}\vspace{-3mm}
	\caption{Restoration results of all methods on band 197 in EO-1 Hyperion data. (a) Original HSI, (b) SVD, (c) RegL1ALM, (d) CWM, (e) MoG-RPCA, \hspace{1cm} (f) PMoEP, (g) LRMR, (h) LRTV, (i) TDL, (j) BM4D, (k) NMoG. \label{EO1197} }
\end{figure*}

\begin{figure*}
	\centering
	\includegraphics[width=0.9\linewidth]{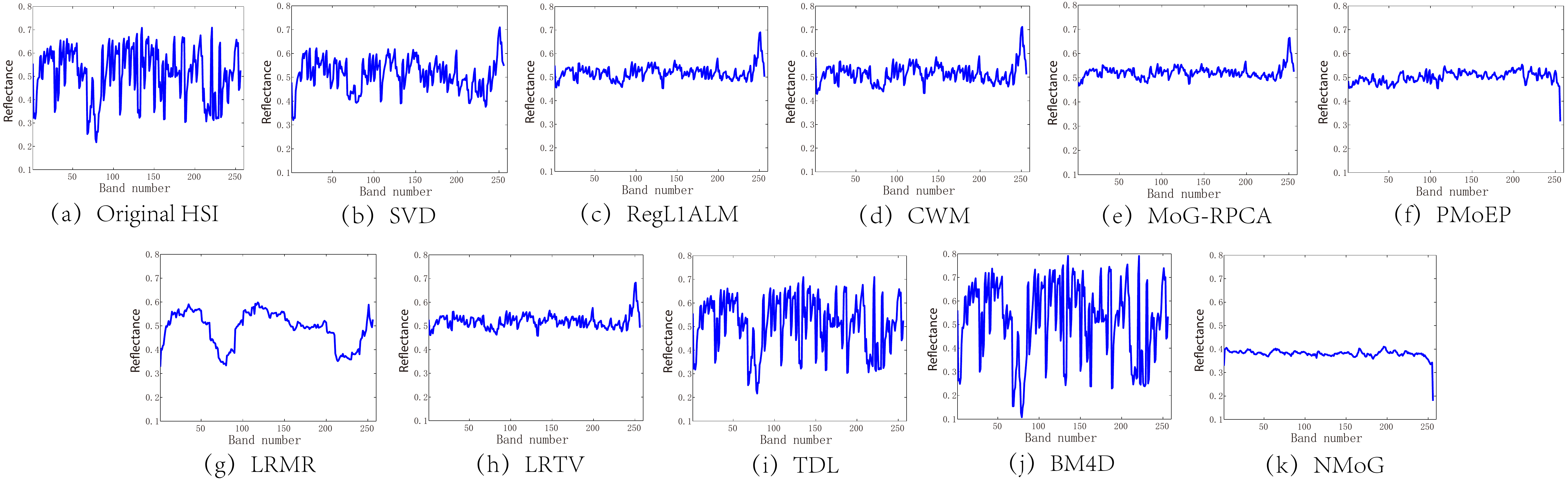}\vspace{-3mm}
	\caption{Vertical mean profiles of band 100 in EO-1 Hyperion data.  (a) Original HSI, (b) SVD, (c) RegL1ALM, (d) CWM, (e) MoG-RPCA,  (f) PMoEP, (g) LRMR, (h) LRTV, (i) TDL, (j) BM4D, (k) NMoG. \label{EO1ver} }
\end{figure*}

The second Hyperion dataset is of the size $ 3128 \times 256 \times 242 $ and some bands are so seriously polluted that all signatures of pixel is equal 0. Thus we only use a subset of its $198$ bands in this experiment after removing those zero signatures bands $ 1 $-$7$, $58$-$76$, and $224$-$242$, which are evidently outliers without any intrinsically useful information. We further spatially cropped a square area from the HSI with size $ 256 \times 256 \times 198 $ with relatively evident noises to specifically testify the denoising capability of a utlized method.

Figs. \ref{EO1100}, \ref{EO1144} and \ref{EO1197} show bands 100, 144 and 197 of the restored HSI images by all competing methods. From the figures, it can be easily observed that SVD and TDL methods have little effects in removing the noise; RegL1ALM and CWM methods can only partially eliminate the noise and MoG-RPCA and PMoEP methods have a better performance than SVD, RegL1ALM and CWM due to their better noise fitting ability brought by the MoG and MoEP models. They, however, still miss lots of details in their HSI recovery due to their implicit improper i.i.d. assumptions on noise distributions under HSI. All other competing methods also cannot achieve a satisfactory HSI denoising results even though they have considered more HSI prior knowledge in their models. Comparatively, the superiority of the proposed method can be easily observed in both detail preservation and noise removing.

Fig. \ref{EO1ver} shows the vertical mean profiles of band 100 before and after restoration. As shown in Fig. \ref{EO1ver} (a), due to the existence of mixed noise especially the stripes disorderly located in the image, there are evident fluctuations over various places of the curve. After the restoration processing, all competing methods show evident non-smoothness over the corresponding curves, except that obtained from the proposed one. This implies that the HSI obtained by proposed NMoG-LRMF method can better preserve such prior knowledge possessed by real HSIs, as also substantiated by Fig. \ref{urbanhor}.

\subsection{Effect of component number on denoising performance}
		\begin{figure}
			\centering
			\includegraphics[width=0.47\linewidth]{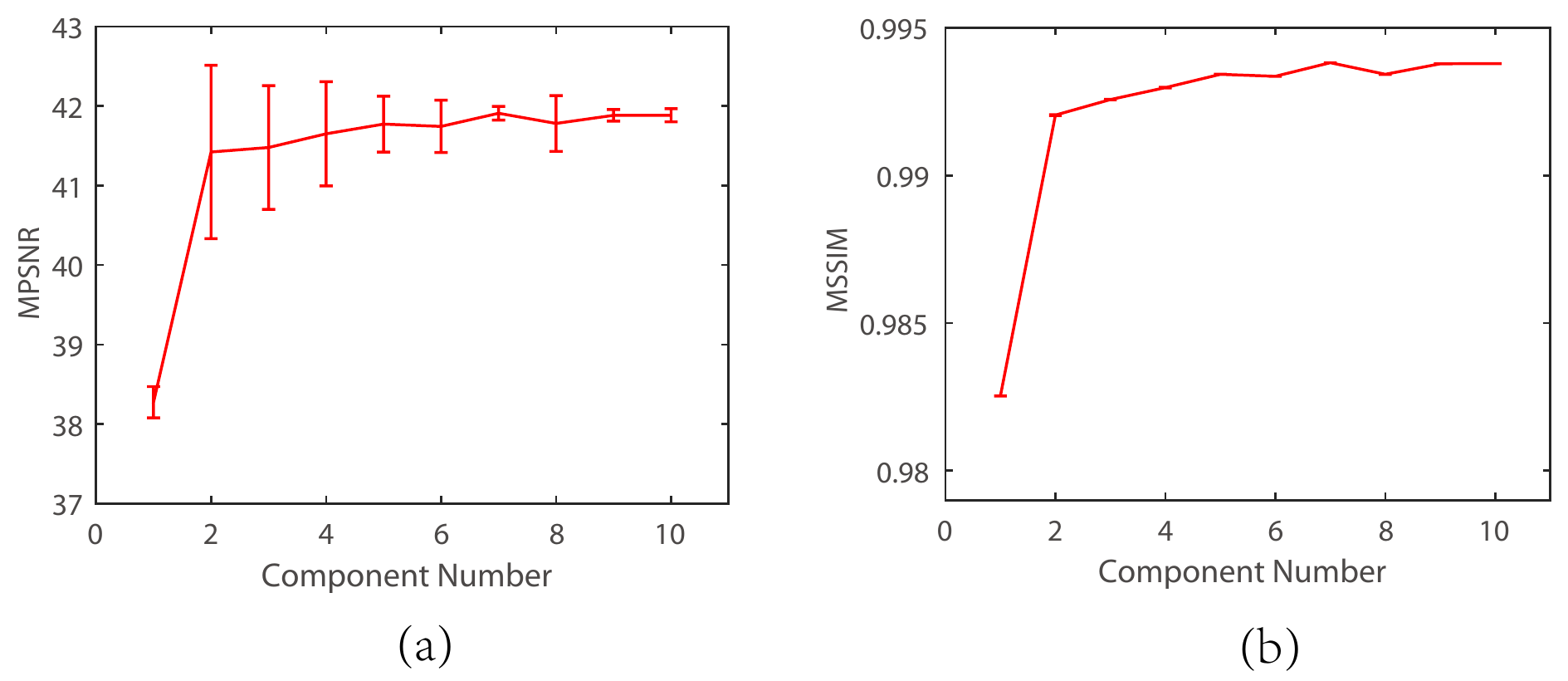}
\includegraphics[width=0.47\linewidth]{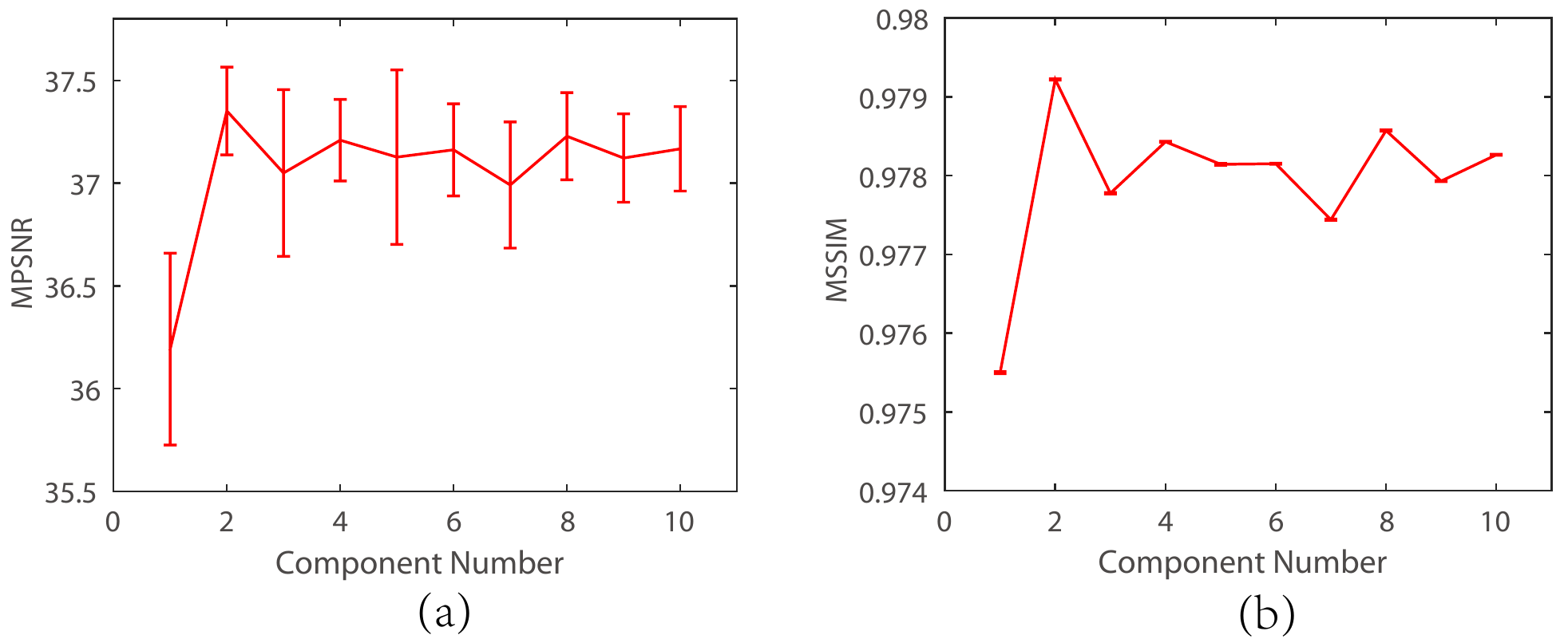}
			\caption{Stability test of the proposed method to Gaussian component number $K$ on DC Mall data in Gaussian $ + $ impluse noise and mixture noise cases. (a) MPSNR tendency curve with respect to $K$.  (b) MSSIM tendency curve with respect to $K$.}	 \label{impluse} 
		\end{figure}
		
In this section, we examine the sensitivity of NMoG-LRMF method to the setting of the Gaussian component number $ K $. We run NMoG-LRMF with 20 initializations on the DC Mall data in Gaussian $ + $ impluse noise and  mixture noise cases, respectively, with $K$ varying from $1$ to $10$. The results are shown in Fig. \ref{impluse}. It can be easily observed that after $K$ is larger than $2$, the denoising performance of the proposed method tends to be stable and not very sensitive to the choise of $K$ value. Actually, in all our real experiments, we just simply set $K$ as $3$, and our method can consistently perform well throughout all our experiments.
				
\section{Conclusion}
In this paper, we initially propose a strategy to model the HSI noise using a non-i.i.d. noise assumption. Then we embed such noise modeling strategy into the low-rank matrix factorization (LRMF) model and propose a non-i.i.d LRMF model under the Bayesian framework. A variational Bayes algorithm
is presented to infer the posterior of the proposed model.
Compared with the current state-of-the-arts techniques, the
proposed method performs more robust due to its capability of
adapting to various noise shapes encountered in applications,
which is substantiated by our experiments implemented on
synthetic and real noisy HSIs.

In our future work, we will try to extend the application of our method to video and face image data. Also, we will integrate more useful HSI prior terms into our model to further enhance its denoising capability. Besides, the proposed noise modeling strategy can be specifically re-designed under certain application context, like the wind speed prediction problem as indicated in~\cite{hu2014noise,hu2016estimating}.

\ifCLASSOPTIONcaptionsoff
  \newpage
\fi







\bibliographystyle{ieee}
\bibliography{bare_jrnl}
\end{document}